\documentclass[letterpaper]{article}
\usepackage{aaai24} 
\usepackage{times} 
\usepackage{helvet} 
\usepackage{courier} 
\usepackage[hyphens]{url} 
\usepackage{graphicx} 
\usepackage{graphicx}  
\usepackage{natbib}  
\usepackage{caption}  
\usepackage[utf8]{inputenc} 
\usepackage[T1]{fontenc}    
\usepackage{booktabs}       
\usepackage{amsmath}
\usepackage{amssymb}
\usepackage{amsthm}
\usepackage{amsfonts}       
\usepackage{nicefrac}       
\usepackage{xcolor}         
\usepackage{mathtools}
\usepackage{pdfpages}

\usepackage{etoolbox}
\newcommand{\insertfig}{\vspace{0.3cm}\includegraphics[width=\textwidth]{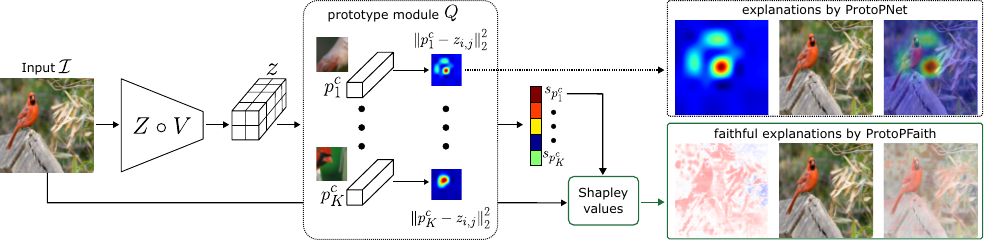}
    \captionof{figure}{In ProtoPNet~\cite{Chen_2019}, an image $\mathcal{I}$ is classified by feeding a vector of distance scores $s_{p_k^c}$ through a linear layer (omitted for clarity).
    Each distance represents the minimum distance between a trainable prototype $p_k^c$ to each spatial latent feature vector in $z$.
    An explanation in ProtoPNet is the up-scaled distance map of a prototype ${\lVert{p^c_k - z_{i,j}}\rVert}^2_2$ overlayed onto the input image.
    In this work, we demonstrate that these explanations do not adhere to established explainability axioms and, thus, cannot be considered faithful representations of the underlying processes. 
    We propose to resolve this shortcoming of ProtoPNet with faithful explanations based on decomposing distance scores in terms of Shapley values w.r.t. the input image.}
    \label{fig:overview}
    \vspace{0.5cm}}
\makeatletter
\apptocmd{\@maketitle}{\centering\insertfig}{}{}
\makeatother
\title{Keep the Faith: Faithful Explanations in Convolutional Neural Networks for Case-Based Reasoning}

\author{
Tom Nuno Wolf\textsuperscript{\rm 1, \rm 2, \rm3},
Fabian Bongratz\textsuperscript{\rm 1, \rm 2, \rm3},
Anne-Marie Rickmann\textsuperscript{\rm1, \rm 2},\\
Sebastian P{\"o}lsterl\textsuperscript{\rm 2},
Christian Wachinger\textsuperscript{\rm 1, \rm 2, \rm3}
}

\newcommand{\ourmod}{ProtoPFaith}
 
\affiliations{
\textsuperscript{\rm 1}Department of Radiology, Technical University Munich, Munich, Germany \\
\textsuperscript{\rm 2}Lab for Artificial Intelligence in Medical Imaging, Ludwig-Maximilians-University, Munich, Germany \\
\textsuperscript{\rm 3}Munich Center for Machine Learning (MCML) \\
tom\_nuno.wolf@tum.de
}

\begin{document}
\maketitle

\begin{abstract}
Explaining predictions of black-box neural networks is crucial when applied to decision-critical tasks.
Thus, attribution maps are commonly used to identify important image regions, despite prior work showing that humans prefer explanations based on similar examples. 
To this end, ProtoPNet learns a set of class-representative feature vectors (prototypes) for case-based reasoning.
During inference, similarities of latent features to prototypes are linearly classified to form predictions and attribution maps are provided to explain the similarity.
In this work, we evaluate whether architectures for case-based reasoning fulfill established axioms required for faithful explanations using the example of ProtoPNet.
We show that such architectures allow the extraction of faithful explanations.
However, we prove that the attribution maps used to explain the similarities violate the axioms.
We propose a new procedure to extract explanations for trained ProtoPNets, named \ourmod.
Conceptually, these explanations are Shapley values, calculated on the similarity scores of each prototype.
They allow to faithfully answer which prototypes are present in an unseen image and quantify each pixel's contribution to that presence, thereby complying with all axioms.
The theoretical violations of ProtoPNet manifest in our experiments on three datasets (CUB-200-2011, Stanford Dogs, RSNA) and five architectures (ConvNet, ResNet, ResNet50, WideResNet50, ResNeXt50).
Our experiments show a qualitative difference between the explanations given by ProtoPNet and \ourmod.
Additionally, we quantify the explanations with the Area Over the Perturbation Curve, on which \ourmod\ outperforms ProtoPNet on all experiments by a factor >$10^3$.
\end{abstract}

\section{Introduction}

With continued progress in deep learning, AI models become ubiquitous in daily life. 
For many tasks, they are able to outperform humans~\cite{shin2023superhuman}. At the same time, it is difficult for humans to understand the decision-making of such complex black-box models with millions of parameters~\cite{adadi2018peeking,arrieta2020explainable}. 
This fundamental issue is particularly relevant in decision-critical areas, such as medicine, finance, or justice.
To increase transparency, methods have been developed that try to locally approximate the decision-making of neural networks in a \emph{post-hoc} fashion~\cite{vale2022explainable}.
However, such explanations can vary dramatically between explanation techniques and models~\cite{Rudin_2019}:
For example, pixel-wise attribution maps aim at highlighting the contribution of each pixel to the output logit, which is realized in~\citeauthor{Simonyan_2013}~\shortcite{Simonyan_2013} via propagation of gradients from the output logit to the input pixels.
Such approaches are directly influenced by the model parameters.
Alternatively, classifying image latents with a $k$-nearest-neighbor (kNN) algorithm allows showing the $k$ most similar images as explanations.
Although the theoretic motivation of pixel-wise attributions is to mimic the visual cortex, prior studies~\cite{jeyakumar2020can,kim2023help,nguyen2021effectiveness} have concluded that example-based explanations are easiest to understand for humans.
While case-based reasoning, as in kNN, answers \emph{what} contributed to a prediction, it does not allow to explain \emph{how} the model transformed the image to arrive at its prediction.
In other words, the explanations give insight into the decision-making of the model but lack to give insight into the high-dimensional non-linear function that a neural network represents.
To compare and analyze explanation techniques from a theoretical point of view, several axioms have been introduced previously~\cite{Lundberg_2017,IntGrad}. 
In this work, we refer to an explanation method that satisfies all axioms as \emph{faithful}.

ProtoPNet~\cite{Chen_2019} is a prominent implementation of case-based reasoning and has been widely adopted, for image classification~\cite{Chen_2019,Donnelly_2022} and decision-critical tasks, like automated diagnosis from chest radiography~\cite{Kim_2021} and dementia-diagnosis~\cite{Wolf_2023}.
It linearly classifies similarities between trainable class-representative feature vectors (\emph{prototypes}) and latent features of unseen images.
Importantly, its inherently interpretable architecture allows the extraction of explanations on a pixel-level.
The model's decision-making is explained by visualizing image crops that each prototype represents next to pixel-level explanations of the unseen image, as seen in Fig.~\ref{fig:overview}, leading to the type of explanations that humans prefer~\cite{jeyakumar2020can,kim2023help,Rudin_2019}.
Therefore, its pixel-level explanations can, supposedly, be incorporated to answer the question \emph{how} it came to the conclusion that the unseen image belongs to a certain class.

In this work, we evaluate whether the explanations given by ProtoPNet are perfectly faithful.
Our theoretical examination shows that case-based reasoning, as implemented by ProtoPNet, i.e., classification based on similarities, allows for faithful explanations.
In contrast, the explanations on the image level assume spatial dependency of latent feature maps and the image space, which breaks in the general case of convolutional neural network (CNN) backbones.
Therefore, we propose to exploit the architectural properties of ProtoPNet to extract attribution maps that faithfully show the contribution of each pixel to the similarities:
We transform a trained ProtoPNet into a lightweight probabilistic model~\cite{Gast_2018} and extract Shapley values with DASP~\cite{Ancona_2019}.
Additionally, we demonstrate that the theoretical violations of ProtoPNet arise in real-world applications. 
We qualitatively show differences in explanations between ProtoPNet and our proposed procedure, named \ourmod, on three established datasets (CUB-200-2011~\cite{CalTechBirds}, Stanford Dogs~\cite{Dogs},  RSNA~\cite{rsna}) and on five CNN backbones (ConvNet, ResNet, ResNet50, Wide-ResNet50, ResNeXt50).
Additionally, we quantify the difference with the Area Over the Perturbation Curve (AOPC)~\cite{Tomsett_2020}.

Our key contributions are as follows:
\begin{itemize}
    \item We prove that the explanations generated by ProtoPNet-like architectures are not faithful to the decision-making of the model.
    \item Instead, we propose to leverage the architecture of case-based reasoning implemented by ProtoPNet to extract explanations based on Shapley values.
    \item \ourmod\ allows efficient extraction of explanations for high-dimensional image inputs and requires conversion of layers into probabilistic layers, for which we derive closed-form solutions for mean and variance if they are not readily available in the literature (i.e., squared L2-norm, ReLU1).
    \item We empirically demonstrate that there are substantial differences, both qualitative and quantitative, between the explanations extracted with ProtoPNet and \ourmod.
    \item Explanations of \ourmod\ outperform ProtoPNet on the AOPC score by a factor >$10^3$ for all experiments.
\end{itemize}

\section{Axiomatic Evaluation of Explanations for Case-Based Reasoning}

First, we evaluate explanations found in the literature with respect to the axioms required for faithful explanations.
Then, we investigate the architecture of case-based reasoning with the example of ProtoPNet and show why some axioms are violated.

\paragraph{Related Work: Pixel-Level Explanation Methods for CNNs.}
Typically, explanations ought to represent each input feature's effect on the prediction. 
\citeauthor{Lundberg_2017}~\shortcite{Lundberg_2017} and~\citeauthor{IntGrad}~\shortcite{IntGrad} independently proposed a set of useful axioms that a deep learning attribution method should fulfill, which can be summarized as:
\begin{itemize}
    \item \textbf{Sensitivity:} If there is a change in the value of a feature and the prediction, the attribution of that feature should not be zero.
    \item \textbf{Implementation Invariance:} Attributions for two models whose predictions are identical for all inputs should be identical.
    \item \textbf{Completeness:} Attributions of two inputs should add up to the difference in the model output.
    \item \textbf{Dummy:} If the prediction of a model is independent of a feature, its attribution should always be zero.
    \item \textbf{Linearity:} If a model $f$ is a linear combination of other models ($f = a f_1 + b f_2$), the attributions of $f$ should follow the same linear combination.
    \item \textbf{Symmetry-Preserving:} If the prediction of a model is identical when replacing two input features with one another, the attribution of both features should be equal.
\end{itemize}

\emph{Gradient-based methods} are characterized by a backward pass of the neural network used to calculate the gradient of a prediction with respect to input features, which can be used to create feature attribution maps~\cite{Simonyan_2013}.
However, as demonstrated by~\citeauthor{IntGrad}~\shortcite{IntGrad}, gradient-based feature attribution methods violate either the \textbf{Sensitivity} axiom due to vanishing gradients in the ReLU activation, concerning methods like~\citeauthor{Baehrens_2010}~\shortcite{Baehrens_2010},~\citeauthor{Simonyan_2013}~\shortcite{Simonyan_2013},~\citeauthor{Springenberg_2014}~\shortcite{Springenberg_2014},~\citeauthor{Zeiler_2014}~\shortcite{Zeiler_2014}, or \textbf{Implementation Invariance} if the gradient is computed in a discrete fashion, as in~\citeauthor{Binder_2016LayerWiseRP}~\shortcite{Binder_2016LayerWiseRP} and~\citeauthor{Shrikumar_2016}~\shortcite{Shrikumar_2016}.
Moreover,~\citeauthor{Shah_2021}~\shortcite{Shah_2021} showed that gradient-based methods tend to suffer from feature leakage, i.e., the contribution of unrelated features to the prediction.
\emph{Activation-based methods}~\cite{Chattopadhay_2018,Desai_2017,Fu_2020,jiang_2021layercam,selvaraju2017grad,Wang_2020score,Zhou_2016}, on the other hand, leverage neuron activations in (usually deep/final) convolutional layers to assess the importance of input regions for the network output. Even though being widely established in practice, these methods typically violate \textbf{Sensitivity} and/or \textbf{Completeness}~\cite{Fu_2020}, which renders their applicability to decision-critical tasks questionable.

Different from gradient- and activation-based methods, \textit{perturbation-based methods} mask or alter an input image feature and calculate the difference to the output of the original input.
Existing work proposed to occlude~\cite{Zeiler_2014}, marginalize with a sliding window~\cite{Zintgraf_2017}, randomly perturb~\cite{Petsiuk}, or occlude parts of an input image with perturbation space-exploration~\cite{fel2022don}.
In~\citeauthor{Dabkowski_2017}~\shortcite{Dabkowski_2017},~\citeauthor{Fong_2017}~\shortcite{Fong_2017}, and~\citeauthor{Ribeiro_2016}~\shortcite{Ribeiro_2016}, the black-box predictor is approximated by an interpretable model locally and super-pixel explanations summarized to the global input.

Another perturbation-based approach is based on \emph{Shapley values}~\cite{Shapley_1953}, which originate from cooperative game theory; each feature is treated as a player in a game and contributes to the final prediction.
Removing a player $i$ from all possible coalitions $S \subseteq P$ of a set of players $P$ , i.e., marginalizing a player, yields its contribution $\psi_i$ to the set function $\hat{f}: P \rightarrow \mathbb{R}$:
\begin{align*}
    \psi_i = \sum_{S \subseteq P \setminus \{i\} }
    \frac{\vert S \vert ! (\vert P \vert - \vert S \vert -1)!}{\vert P \vert !}
    \left(\hat{f}(S \cup \{i\}) - \hat{f}(S)\right).
\end{align*}
In contrast to other attribution methods, Shapley values satisfy \emph{all} axioms and, thus, provide faithful explanations~\cite{Covert_2020,Lundberg_2017,IntGrad,Zheng_2022}.
Specifically, they satisfy \textbf{Completeness}, as the sum of contributions of all players equals the prediction:
\begin{align}
    \sum_i \psi_i = \hat{f}(P) - \hat{f}(\emptyset).
\end{align}
When aiming to compute Shapley values for an input image $\mathcal{I}$ of a DNN, $\hat{f}(S)$ signifies the output of the DNN when all pixels not in $S$ are replaced by a baseline value.
Approximating Shapley values has gained a lot of attention~\cite{Ancona_2019,Lundberg_2017,Shrikumar_2017,Strumbelj2014explaining,IntGrad,wang2022accelerating}, as their exact calculation grows exponentially with the number of input features, with DASP~\cite{Ancona_2019} outperforming all other approximation methods in terms of approximation error.

\paragraph{Preliminary: Interpretability with ProtoPNet-like Architectures.}

ProtoPNet, outlined in Fig.~\ref{fig:overview}, implements case-based reasoning as a function $f(\mathcal{I}) = (F \circ Q \circ Z \circ V)(\mathcal{I})$ mapping an input image $\mathcal{I} \in \mathbb{R}^{H \times W \times \tilde{C}}$ to a set of output logits.
More precisely, a CNN encoder $V: \mathbb{R}^{H \times W \times \tilde{C}} \rightarrow \mathbb{R}^{H^{\prime} \times W^{\prime} \times C^{\prime}}$, extract features and is typically pre-trained on the desired task ($H$, $W$, $\tilde{C}$ the spatial height, width, and channel dimension of the input image; $H^{\prime}$, $W^{\prime}$, $C^{\prime}$ the latent feature map height, width, and channel dimensions).
The feature extractor $Z: \mathbb{R}^{ H^{\prime} \times W^{\prime} \times C^{\prime}} \rightarrow \mathbb{R}^{H^{\prime} \times W^{\prime} \times L}$ maps to a latent feature map $z$, which matches the channel-size $L$ of prototypes, and consists of two $1 \times 1$ convolutions and non-linearities.
The prototype module $Q: \mathbb{R}^{H^{\prime}\times W^{\prime} \times L} \rightarrow \mathbb{R}^{K \cdot C}$ extracts a distance vector $s$ for $K$ prototypes per class $C$.
It consists of the minimum distances $s_{p^c_k}$ of the squared L2-norm of each prototype to all spatial latent feature vectors $z_{i,j}$ in $z$:
\begin{equation}
\begin{aligned}\label{formula:sim}
    s_{p_k^c}(\mathcal{I}) &= \min_{i = 1,\dots, H^{\prime},j = 1, \dots, W^{\prime}}~{\lVert{p^c_k - z_{i,j}}\rVert}^2_2
    \\ &= \min_{i = 1,\dots, H^{\prime},j = 1, \dots, W^{\prime}}
    \sum_{l=1}^L \left(p^c_{k,l} - z_{i,j,l}\right)^2,
\end{aligned}
\end{equation}
with $p^c_k \in \mathbb{R}^{1 \times 1 \times L}$ and $z = (Z \circ V)(\mathcal{I}) \in \mathbb{R}^{H^{\prime} \times W^{\prime} \times L}$.
Lastly, the classification layer $F: \mathbb{R}^{K \cdot C} \rightarrow \mathbb{R}^{C}$ is implemented with a single linear layer.

Importantly, prototypes $p^c_k$ are trainable parameters (vectors) of the network.
After a certain number of iterations, each prototype is replaced by the closest (squared L2-distance) latent feature vector $z_{i,j}$ extracted from all samples of the training set that are of class $c$.
Thus, a prototype $p^c_k$ represents exactly one class-representative latent feature vector $z_{i,j}$ from a training image.
An unseen image is classified by feeding the distance vector $s$ (see Fig.~\ref{fig:overview}) to the classification layer $F$.

The \emph{distance map} ${\lVert{p^c_k - z_{i,j}}\rVert}^2_2$ is utilized to extract pixel-level explanations.
First, the maximum distance of the distance map is selected, which is globally defined for all possible inputs if the last layer of the feature extractor $Z$ is a bounded non-linearity.
The maximum distance is subtracted by the values of the distance map\footnote{Originally,~\citeauthor{Chen_2019}~\shortcite{Chen_2019} used a log-activation instead, which we discuss in detail in the next section.}.
Then, the flipped distance map is up-scaled to the input image size and overlaid with the input image, forming an \emph{attribution map}.
A prototype is visualized by extracting the image crop around the 95\%-percentile region of the corresponding attribution map.

\paragraph{Theoretic Evaluation of the Faithfulness of ProtoPNet Explanations.}\label{sec:ppnetbad}

The decision-making of ProtoPNet is modeled as a linear function $F$, which maps distances $s_{p^c_k}$ to class logits.
A linear mapping satisfies all introduced axioms~\cite{Strumbelj2014explaining}.
Therefore, the model prediction, i.e., the decision-making of the model, is faithful with respect to the distance vector $s$.
However, ProtoPNet adds a log-activation to the prototype module,  i.e., $Q^{\prime}(s_{p_k^c}(\mathcal{I}))=\log((s_{p_k^c}(\mathcal{I}) + 1) / (s_{p_k^c}(\mathcal{I}) + \epsilon_{\rightarrow 0}))$.
This introduces non-linearity between the distance maps (which are used to yield attribution maps) and the classification.
Therefore, the \textbf{Linearity} axiom is not preserved.

\begin{figure}
    \centering
    \includegraphics[width=\columnwidth]{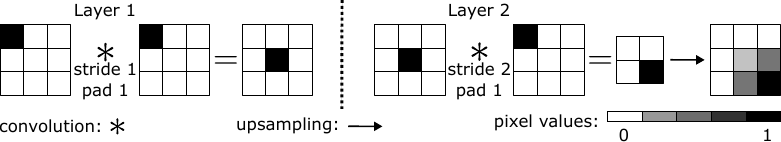}
    \caption{An up-sampled latent feature map itself cannot serve as a spatial indicator of pixel attribution in the general case.}
    \label{fig:counterproof}
\end{figure}

As explained in the previous section, attribution maps are the result of up-scaling distance maps from the latent space to the image space.
This process implies a spatial relationship between distances in latent space and image space, as the distances of prototypes to latent feature vectors are calculated over the spatial dimensions in the latent space.
In established CNN architectures used in ProtoPNet (VGG~\cite{Simonyan_2015}, ResNet~\cite{He_2015}, DenseNet~\cite{Huang_2017}), the locality of latent features is, however, lost after a few layers, i.e., the receptive field of each latent feature is the whole input image.
We now prove by counterexample that there is no general spatial dependency between latent feature maps and the input image space:
\begin{proof}
Suppose a CNN consists of two convolutional layers (one input and output channel with linear activation) with kernel weights $\theta_1 \in \{0,1\}^{3 \times 3 \times 1}$ and $\theta_2 \in \{0, 1\}^{3 \times 3 \times 1}$, as depicted in Fig.~\ref{fig:counterproof}.
Suppose the first layer has a stride and padding of one, and the second layer has a stride of two and padding of one. 
Both kernel weights are identical, with zeros everywhere beside the top left value, which is one.
Feeding an exemplary input of size $3 \times 3$, consisting of zeros with the exception of the top left feature set to one, to this network yields an output of size $2 \times 2$, which consists of zeros and the bottom right feature activated at one.
Upscaling this latent feature map to the input dimension, as done in ProtoPNet, creates an attribution map that suggests the bottom right part of the input to be relevant.
However, the only feature that activated the bottom right latent feature is the top left feature in the input space.
Thus, there is no spatial relation between the features of the input space and the latent space.
\end{proof}
As a result, features that do not affect the latent features, and should therefore be treated as dummies, have a high attribution.
Therefore, the \textbf{Dummy} axiom is not fulfilled.

Additionally, ProtoPNet explanations violate the axiom \textbf{Completeness}, as each distance $s_{p_k^c}$ is exactly the distance of \emph{one} latent feature vector of the latent feature map, i.e., $\min_{i = 0,\dots, H^{\prime},j = 0, \dots, W^{\prime}}~{\lVert{p^c_k - z_{i,j}}\rVert}^2_2$.
Therefore, the sum of pixel-level attributions does not equal the distance.
For the same reason, the attribution maps violate the \textbf{Dummy} axiom.
Lastly, thresholding the attribution map with the 95\%-percentile violates \textbf{Sensitivity}:
As the percentile is chosen heuristically, it suppresses the attributions of some input features that have non-zero attribution.
In contrast, we will show that \ourmod\ calculates attribution maps with respect to the minimum distance $s_{p_k^c}$ faithfully, which manifests in our experiments.

In summary, the decision-making of ProtoPNet requires a linear activation to be faithful, while the explanations provided on an image-level break the overall faithfulness of the model, as they break \textbf{Completeness}, \textbf{Dummy}, and \textbf{Sensitivity}.
Next, we introduce \ourmod\ to extract faithful explanations from ProtoPNet, rendering it feasible for decision-critical tasks.

\section{Methods}

As described in the previous section, the explanations given by ProtoPNet are only faithful if \textbf{Linearity} of the classification w.r.t. minimum distances $s_{p_k^c}$ can be restored, and attribution maps are faithful to changes in the minimum distances themselves.
We showed that the attribution maps are not faithful.
In contrast, Shapley values are guaranteed to satisfy \emph{all} required axioms, albeit their exact calculation requires exponentially many model evaluations.
To this end, DASP~\cite{Ancona_2019} has shown to approximate true Shapley values with only few model evaluations, rendering it feasible for high-dimensional image inputs while outperforming competing methods in terms of approximation error.
We show below how to convert ProtoPNet with this framework.
Our proposed method to extract explanations, named \ourmod, is an important step towards a transparent ProtoPNet, allowing to faithfully explain the case-based decision-making referenced by~\citeauthor{Chen_2019}~\shortcite{Chen_2019} as "\emph{this} looks like \emph{that}".
The source code is available at \url{https://github.com/ai-med/KeepTheFaith}.

\paragraph{Leveraging Faithful Explanations in ProtoPNet.}

We propose to restore \textbf{Linearity} of the classification of the distance vector $s$ by dropping the log-activation introduced by ProtoPNet (note that the minimum of the squared L2-norm is zero).
Instead of showing up-scaled log-activations of distance maps and the related weight of the classification layer, we need to calculate a contribution score $\Psi_k$ of a similarity to the log-probability:
{
\allowdisplaybreaks
\begin{align*}
    & \log(P(y=c~|~\mathcal{I})) \\
    & = \log\left(\frac{\exp{\sum_{k=1}^K -a_{c,k} s_{p_k^c}(\mathcal{I})}}{\sum_{\hat{c}=1}^C \exp{\sum_{k=1}^K -a_{\hat{c},k} s_{p_k^{\hat{c}}}(\mathcal{I})}} \right) \\
    & = \log\left(\frac{\exp{\sum_{k=1}^K -a_{c,k} s_{p_k^c}(\mathcal{I})}}{R} \right) \\
    & = \sum_{k=1}^K -a_{c,k} s_{p_k^c}(\mathcal{I}) - \log{R} \\
    & = \sum_{k=1}^K \Psi_k\text{, with }\Psi_k  = -a_{c,k} s_{p_k^c}(\mathcal{I}) - \frac{\log{R}}{K},
\end{align*}
}
where $a_{c,k}$ are the coefficients of the linear layer $F$ and $y$ is the target label of the image $\mathcal{I}$.
Note that we use the contribution score to compare the contribution of prototypes of the desired class, i.e. high contribution scores contribute more than low contribution scores.

In ProtoPNet, we are interested in the portion of an image that is representative of a prototype.
Therefore, we propose to utilize the implementation for case-based reasoning introduced by ProtoPNet to extract Shapley values via the minimum distance $s_{p_k^c}$.
After network training, precisely one image exists in the training set from which a prototype arises.
As a result, we can visualize a prototype faithfully by calculating the Shapley values w.r.t. the distance between this training image and the prototype (distance is 0 for this image).
The resulting explanations highlight the pixels that are compressed into a prototype.

Likewise, we calculate the Shapley values of pixels w.r.t. the squared L2-distance of a prototype and visualize them for unseen test images.
As the vector of minimum distances $s$ is fed through the final linear layer only, we can use the \textbf{Linearity} axiom to transform similarity-based Shapley values into attribution w.r.t. the model prediction.
This would be the weighted sum of all attribution maps.
To allow quantitative comparison between pixel attributions of different models and prototypes, we opt for a bounded non-linearity before calculating the distances.
Furthermore, a bounded non-linearity, such as ReLU1 (Eq.~\ref{eq:relu1}), helps for faster model convergence and allows us to derive the first and second-order moments analytically, which is required for DASP.

\paragraph{Approximation of Shapley Values with DASP.}

Recently, advances in uncertainty propagation for DNNs~\cite{Gast_2018} were applied to approximate Shapely values in DASP~\cite{Ancona_2019} with a few network evaluations.
This is desirable when the number of input features is high, enabling efficient approximation of Shapley values of a whole input image.

DASP calculates the contribution $\psi_{i,d}$ of the $i$-th feature to a coalition $S_d$ of fixed size $d$ by modeling each $S_d$ as an aleatoric uncertainty, which is propagated through a probabilistic network to yield an expected value $\mathbb{E}_d[\psi_{i,d}]$.
Then, the expectation of a Shapley value is estimated as:
\begin{align}
    \mathbb{E}[\psi_i] = \frac{1}{\vert P \vert}\sum_{d=0}^{\vert P \vert - 1}\mathbb{E}_d[\psi_{i,d}].
\end{align}
If a DNN can be converted into a probabilistic model, each $\mathbb{E}_d[\psi_{i,d}]$ can be calculated with a single forward pass.
To this end, closed-form solutions for mean and variance need to be derived.
While they were summarized for standard DNN layers in~\citeauthor{Ancona_2019}~\shortcite{Ancona_2019}, they are not yet readily available for the prototype module $Q$ and the ReLU1 non-linearity.
Hence, we derive them in the following.
Notably, the converted probabilistic model does not need to be trained, as the trained weights of a ProtoPNet are re-used.

\paragraph{Derivation of Lightweight Probabilistic Layers for ProtoPNet.}

We analytically derive expectation $\mathbb{E}$ and variance $\mathbb{V}$ of a layer input mean $\mu$ and variance $\sigma^2$ for ReLU1 for $\mathbb{E}(\mu, \sigma) = \int g(x) \frac{1}{\sigma} \phi(\frac{x - \mu}{\sigma})\,dx$ and variance $\mathbb{V}(\mu, \sigma) = \int (g(x) - \mathbb{E}(\mu, \sigma))^2 \frac{1}{\sigma} \phi(\frac{x - \mu}{\sigma})\,dx$, with $\phi$ the standard Gaussian probability density function $\phi (x) = \frac{1}{\sqrt{2\pi}}e^{-\frac{x^2}{2}}$, the corresponding cumulative distribution function $\Phi(x) = \int_{-\infty}^{x}  \phi(t)\,dt$~\cite{Frey_1999}, and $g(x) = \text{ReLU1}(x)$ defined as:
\begin{equation}\label{eq:relu1}
 \text{ReLU1}(x) =
 \begin{cases*}
    0, & if x $<$ 0, \\
    x, & if 0 $\leq$ x $\leq$ 1, \\
    1, & if x $>$ 1.
 \end{cases*}
\end{equation}
The expectation and variance are (full derivation in Sec.~A.1, which can easily be extended to any bound other than 1):
\begin{equation}
\begin{aligned}
& \mathbb{E} \left(\mu, \sigma\right) =: \Bar{\mu} = \sigma\left(\phi\left(-\frac{\mu}{\sigma}\right) - \phi\left(\frac{1-\mu}{\sigma}\right)\right) \\ & + \mu\left(\Phi\left(\frac{1-\mu}{\sigma}\right) - \Phi\left(-\frac{\mu}{\sigma}\right)\right) + 1 - \Phi\left(\frac{1-\mu}{\sigma}\right) 
\end{aligned}
\end{equation}
\begin{equation}
\begin{aligned}
    \mathbb{V}\left(\mu, \sigma\right) =~ & \left(\mu^2 - 2 \mu \Bar{\mu} + \sigma^2 + 2 \Bar{\mu} - 1\right) \Phi\left(\frac{1-\mu}{\sigma}\right) \\ &- \left(\mu^2 - 2 \mu \Bar{\mu} + \sigma^2\right) \Phi\left(-\frac{\mu}{\sigma}\right)
    \\ & - \left(\mu \sigma - 2 \Bar{\mu} \sigma + \sigma\right) \phi\left(\frac{1-\mu}{\sigma}\right)
    \\ & + \left(\mu\sigma - 2 \Bar{\mu}\sigma\right) \phi\left(-\frac{\mu}{\sigma}\right)
    \\ & + \Bar{\mu}^2 - 2 \Bar{\mu} + 1
\end{aligned}
\end{equation}
Further, we derive expectation and variance of $s_{p_k^c}$ (see Eq.~\ref{formula:sim}) as follows:
As seen in~\citeauthor{Ancona_2019}~\shortcite{Ancona_2019}, we assume independence of the input Gaussian signals $z_{i,j,l}$.
Thus, we can represent each latent feature vector $z_{i,j}$ as a multivariate Gaussian with diagonal co-variance matrix $\boldsymbol{\Sigma}_{i,j} \in \mathbb{R}^{L \times L}$ consisting of the individual means $\mu_{i,j,l}$ and variances $\sigma^2_{i,j,l}$, i.e., $z_{i,j} \sim \mathcal{N}(\boldsymbol{\mu}_{i,j}, \boldsymbol{\Sigma}_{i,j})$, $\boldsymbol{\mu}_{i,j} \in \mathbb{R}^{L}$.
We can then introduce a new multivariate random variable $Y_{i,j}$:
\begin{align*}
    Y_{i,j} = z_{i,j} - p^c_k \quad \rightarrow \quad Y_{i,j} \sim \mathcal{N}(\boldsymbol{\mu}_{i,j} - p^c_k, \boldsymbol{\Sigma}_{i,j}).
\end{align*}
Now, we calculate the squared L2-norm in Eq.~\ref{formula:sim} as $Z_{i,j} = \sum_{l=1}^L Y_{i,j,l}^2 = Y_{i,j}^T Y_{i,j}$, which allows calculating mean and variance via quadratic forms of random variables~\cite{bruno}:
\begin{align}
    & \mathbb{E}[Z_{i,j}] = \mathbf{trace}(\boldsymbol{\Sigma}_{i,j}) + Y_{i,j}^T Y_{i,j} = \tilde{\mu}_{i,j} \\
    & \mathbb{V}[Z_{i,j}] = 2~\mathbf{trace}(\boldsymbol{\Sigma}_{i,j}^2) + 4 \tilde{\mu}_{i,j}^T \boldsymbol{\Sigma}_{i,j} \tilde{\mu}_{i,j} = \tilde{\sigma}_{i,j} \\
    & Z_{i,j} \sim \mathcal{N}(\tilde{\mu}_{i,j}, \tilde{\sigma}_{i,j}^2).
\end{align}
Finally, we extract the minimum distance with max-pooling over all negated $Z_{i,j}, i=1,\dots,H^{\prime}, j=1,\dots,W^{\prime}$, for which mean and variance are given in~\citeauthor{Ancona_2019}~\shortcite{Ancona_2019}.

\section{Experiments and Results}

We evaluate \ourmod\ on datasets similar to the ones used by ProtoPNet and its adaptations~\cite{Chen_2019,Donnelly_2022,Kim_2021}: CUB-200-2011~\cite{CalTechBirds}, Stanford Dogs~\cite{Dogs}, and a subset of RSNA\footnote{available at https://www.rsna.org/education/ai-resources-and-training/ai-image-challenge/RSNA-Pneumonia-Detection-Challenge-2018}~\cite{rsna} consisting of pneumonia and healthy samples only. 
Tab.~\ref{tab:clf} reports the classification performance of individual experiments, Sec.~A.2 lists training details, and Sec.~A.3 and Sec.~A.4 present additional results.

\begin{table*}[ht]
\centering
\begin{tabular}{ll|ll|ll}
\toprule
      Dataset &         Model & Validation BAcc $\uparrow$ &      Test BAcc $\uparrow$   &       $\text{AOPC}_{\text{ProtoPNet}}$ $\downarrow$    & $\text{AOPC}_\text{{\ourmod}}$ $\downarrow$ \\
\midrule
 CUB-200-2011 &     ResNeXt50 &  72.8 $\pm$ 0.5 & 72.0 $\pm$ 0.6  & -0.000446 & -3.721184  \\
Stanford Dogs &     ResNeXt50 &  84.1 $\pm$ 0.3 & 83.4 $\pm$ 0.6  & -0.001767  &  -1.850909 \\
         RSNA &       ConvNet &  74.0 $\pm$ 10.1 & 72.6 $\pm$ 10.8  & -0.001360 & -8.816450\\
\bottomrule
\end{tabular}
\caption{Balanced Accuracy (BAcc) and AOPC scores for models visualized in Fig.~\ref{fig:2example}. See Tab.~A1 and Tab.~A2 for full results.
}
\label{tab:clf}
\end{table*}

\begin{figure*}[!ht]
    \centering
    \includegraphics[width=\textwidth]{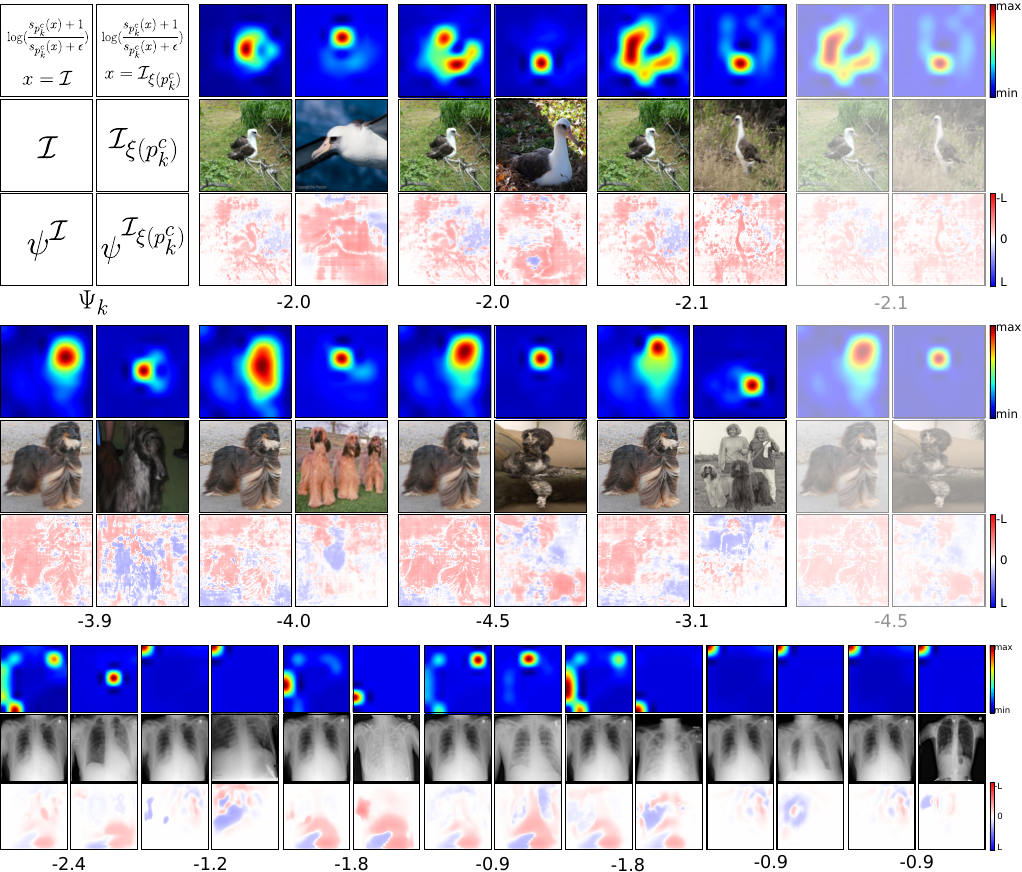}
    \caption{Explanations of the forward pass of: (top) a bird with ResNeXt trained on CUB-200-2011; (middle) a dog with ResNeXt trained on Stanford Dogs; (bottom) ConvNet trained for pneumonia detection on RSNA. We explain the layout of each explanation in the top left with formal notation (left to right, top to bottom): (1) explanation of ProtoPNet for the occurrence of a prototype within the test image; (2) explanation of ProtoPNet for the activation of an image, from which a prototype was extracted; (3) test image; (4) training image, from which the prototype was extracted; (5) explanation of~\ourmod\ for the occurrence of a prototype within the test image; (6) explanation of~\ourmod\ about the image, from which the prototype was extracted. We grayed out duplicate prototypes, which are identical feature vectors. $\Psi_k$ denotes the prototype contribution towards log-probability. See Sec.~A.4 for more visual results.\label{fig:2example}}
\end{figure*}

\paragraph{Qualitative Evaluation.}
Fig.~\ref{fig:2example} visualizes explanations of image predictions from the test sets.
Each predicted image was classified correctly by the model.
Therefore, we visualize the attributions of the prototypes of that class and the corresponding contribution scores $\Psi_k$, which allows us to evaluate which prototype contributed most to the prediction (greater $\Psi_k$).

The model learned duplicate prototypes when trained on CUB-200-2011 and Stanford Dogs.
As seen in almost all explanations of prototypes, attribution maps of ProtoPNet appear focused on a small location of an image.
Its prototype activations for the test image are less sparse, except for CUB-200-2011, in which some attributions comprise background, while the prototype appeared to capture the animal only.
There are no duplicate prototypes for the pneumonia class on the RSNA dataset.
However, almost all prototype explanations of ProtoPNet are located in the background of their corresponding image.
In contrast, the prototypes found by \ourmod\ comprise the body and head for most prototypes of birds.
For dogs, the attributions of prototypes are not as apparent, but always focus on the hair of the animal around the face.
Additionally, the explanations given by \ourmod\ contain large portions of the background.
While the log-activations of ProtoPNet are not helpful for pneumonia on the RSNA dataset, explanations of \ourmod\ capture parts of the lung, with healthy parts indicating a dissimilarity to the prototype.
However, all prototypes in RSNA encode general anatomy like the heart and spine.
Attributions of the test image are similar for most prototypes.
Additional results are presented in Sec.~A.3.

\paragraph{Quantitative Evaluation.}\label{aopc}

We evaluate the explanations given by ProtoPNet and extracted with \ourmod\ based on the squared L2-distance $s_{p_k^c}$.
For each prototype of a model, we select the input image that this prototype originates from (note that per definition of the training procedure, the squared L2-distance $s_{p_k^c}$ between each prototype and its input image equals 0).
We iteratively remove the most relevant features (according to the attribution map of this prototype), as proposed in~\citeauthor{Tomsett_2020}~\shortcite{Tomsett_2020} as the Area Over the Perturbation Curve (AOPC):

\begin{equation}
\begin{aligned}\label{eq:aopc}
    & \text{AOPC}(Q \circ Z \circ V) =~\\ & \frac{1}{C + K + T - 1} \sum^{C}_{c=1}\sum^{K}_{k=1}\sum^{T}_{t=1} s_{p^{c}_k}(\mathcal{I}_{\xi{(p_k^{c})}}^{(0)}) - s_{p^{c}_k}(\mathcal{I}_{\xi{(p_k^{c})}}^{(t)}),
\end{aligned}
\end{equation}
where $\xi{(p_k^{c})}$ denotes a mapping from prototypes to the corresponding index of an image in the training set, and $^{(t)}$ indicates the $t$ most important features removed.
It is expected that removing the most important features first leads to a faster decrease in the AOPC (negative values) if explanations are more meaningful.
Tab.~\ref{tab:clf} demonstrates that removing features according to \ourmod\ yields a decrease in AOPC that differs in several orders of magnitude from removing features according to ProtoPNet.
Thus, the explanations given by \ourmod\ are more accurate and discriminative than the original explanations given by ProtoPNet.

\section{Discussion}

As seen in Fig.~\ref{fig:2example}, the explanations of ProtoPNet and \ourmod\ are very different.
While the explanations of ProtoPNet appear sparse, the close approximation of Shapley values given by \ourmod\ indicates that prototypes learned by the network indeed comprise most of the input domain.
The introduction of granularity on a pixel-level prohibits extraction of crops of the input image that are believed to represent a prototype by ProtoPNet.
While the proposed explanations are harder to interpret, they are faithful to $how$ the model transformed the input into a prototype and $what$ lead to the classification.
For medical applications like pneumonia detection, granularity can be beneficial: When the actual disease is behind ribs, attributions of \ourmod\ do not comprise the ribs, as seen in Fig.~\ref{fig:2example}.
The background explanations found by ProtoPNet would not allow a clinician to learn anything about the model's decision-making.
Enlightened by theoretical derivation and the real-world explanations of \ourmod, we can infer that the features encoded in the prototype are indeed just mapped to an arbitrary position in latent space.
During training of ResNet on RSNA (see Fig.~A1), the prototypes seemed to collapse for each class.
This model achieved a test BAcc of 79.8\% nevertheless.
This is possible because the loss introduced by~\citeauthor{Chen_2019}~\shortcite{Chen_2019} does not enforce prototypes of a single class to be distant from one another.
With a typical black-box, identifying such unexpected issues would be impossible.
Finally, human-understandable explanations, as given by ProtoPNet, are typically designed to fulfill a desired property like sparsity or locality.
However, opting for transparency as in \ourmod\ involves a trade-off, which likely applies to any explainability method.
Hence, future work needs to address the question of how to accomplish explanations that are both human-understandable \emph{and} faithful.

\paragraph{Limitations}

\ourmod\ is currently only applied to single-label image classification but can be extended to multi-label classification as done by XProtoNet~\cite{Kim_2021}, in which prototypes only contribute to the prediction of their own class.
This allows the calculation of Shapley values w.r.t. the output logit, yielding attribution maps in image space for each output logit.
Using this attribution map, we would be able to infer the attribution map of each prototype via using the \textbf{Linearity} property reversely, which would speed up the computation of attribution maps (see Sec.~A.2).

\section{Conclusion}
We elucidated the conceptual and theoretical problems of explanations for case-based reasoning with ProtoPNet.
While this work focuses on the improvement of the transparency of ProtoPNet, it can easily be applied to similar case-based reasoning architectures such as ProtoTrees~\cite{nauta2021neural} or XProtoNet~\cite{Kim_2021}.
The results highlighted that explanations given by ProtoPNet are merely well-defined rather than giving insight into $how$ the model transformed an image.
We showed that the theoretical flaws can be overcome by estimating Shapley values of the similarity score with minor adaptations of the architecture.
Quantitatively, the explanations of \ourmod\ outperformed the explanations of ProtoPNet on the AOPC by a factor >$10^3$ in all experiments.

\section*{Acknowledgements}
This research was partially funded by the Deutsche Forschungsgemeinschaft (DFG, German Research Foundation).
The authors gratefully acknowledge the computational and data resources provided by the Leibniz Supercomputing Centre (www.lrz.de).

{
\small
\bibliography{main}
}
\end{document}


\maketitle
\appendix

\renewcommand{\thetable}{\Alph{section}\arabic{table}}
\renewcommand{\thefigure}{\Alph{section}\arabic{figure}}
\section{Appendix}

\subsection{Derivation of ReLU1 Moments}\label{sec:closedforms}

We analytically derive expectation $E$ and variance $V$ of a layer input mean $\mu$ and variance $\sigma^2$ for ReLU1 for $E(\mu, \sigma) = \int g(x) \frac{1}{\sigma} \phi(\frac{x - \mu}{\sigma})\,dx$ and variance $V(\mu, \sigma) = \int (g(x) - E(\mu, \sigma))^2 \frac{1}{\sigma} \phi(\frac{x - \mu}{\sigma})\,dx$, with $\phi$ the standard Gaussian probability density function $\phi = \frac{1}{\sqrt{2\pi}}e^{-\frac{x^2}{2}}$, the corresponding cumulative distribution function $\Phi(x) = \int_{-\infty}^{x}  \phi(t)\,dt$~\citep*{Frey_1999}, and $g(x) = \text{ReLU1}(x)$ defined as:
\begin{equation*}
 \text{ReLU1}(x) =
 \begin{cases*}
    0, & if x $<$ 0, \\
    x, & if 0 $\leq$ x $\leq$ 1, \\
    1, & if x $>$ 1.
 \end{cases*}
\end{equation*}

Note that this derivation can serve as a blueprint to derive mean and variance for any ReLU with an upper bound by adapting the integration limits.
We use color coding to emphasize where replaced equations come from.

\begin{align*}\label{der:E_ReLU1}
    E(\mu, \sigma) = & \int_{-\infty}^{\infty} \text{ReLU1}(x) \frac{1}{\sigma} \phi(\frac{x - \mu}{\sigma})\,dx \\ 
    = & \int_{-\infty}^{0} \text{ReLU1}(x) \frac{1}{\sigma} \phi(\frac{x - \mu}{\sigma})\,dx + \int_{0}^{1} \text{ReLU1}(x) \frac{1}{\sigma} \phi(\frac{x - \mu}{\sigma})\,dx  + \int_{1}^{\infty} \text{ReLU1}(x) \frac{1}{\sigma} \phi(\frac{x - \mu}{\sigma})\,dx \\
    = & {\color{red}{}\int_{0}^{1} \frac{x}{\sigma} \phi(\frac{x - \mu}{\sigma})\,dx} + {\color{blue}{}\int_{1}^{\infty} \frac{1}{\sigma} \phi(\frac{x - \mu}{\sigma})\,dx}.
\end{align*}
\hrule
\begin{align*}
    {\color{red}{}\int_{0}^{1} \frac{x}{\sigma}} & {\color{red}{}\phi(\frac{x - \mu}{\sigma})\,dx} = \frac{1}{\sigma} \int_{0}^{1} x \phi(a+bx)\,dx, \text{ with }a = -\frac{\mu}{\sigma} \text{, }b = \frac{1}{\sigma} \\ 
     & \stackrel{\text{Eq.101~\citep*{Owen_1980}}}{=} \frac{1}{\sigma} [\left.-\frac{1}{b^2}(\phi(a + bx) + a\Phi(a+bx))]\right\vert_{0}^{1} \\
    = &\frac{1}{\sigma} [\left.-\sigma^2(\phi(\frac{x-\mu}{\sigma}) - \frac{\mu}{\sigma}\Phi(\frac{x-\mu}{\sigma}))]\right\vert_{0}^{1} \\
    = & \left.-\sigma\phi(\frac{x-\mu}{\sigma}) + \mu\Phi(\frac{x-\mu}{\sigma})\right\vert_{0}^{1} \\
    = & -\sigma\phi(\frac{1-\mu}{\sigma}) + \mu\Phi(\frac{1-\mu}{\sigma}) + \sigma\phi(-\frac{\mu}{\sigma}) - \mu\Phi(-\frac{\mu}{\sigma}) \\
    = & ~{\color{red}{}\sigma(\phi(-\frac{\mu}{\sigma}) - \phi(\frac{1-\mu}{\sigma})) + \color{red}{}\mu(\Phi(\frac{1-\mu}{\sigma}) - \Phi(-\frac{\mu}{\sigma}))}.
\end{align*}
\hrule
\begin{align*}
    {\color{blue}{}\int_{1}^{\infty} \frac{1}{\sigma} \phi(\frac{x - \mu}{\sigma})\,dx} = & \left.\Phi(\frac{x-\mu}{\sigma})\right\vert_{1}^{\infty} \\
    = & \left.1 - \Phi(\frac{x-\mu}{\sigma})\right\vert_{-\infty}^{1} \\
    = & ~{\color{blue}{}1 - \Phi(\frac{1-\mu}{\sigma})}.
\end{align*}
\hrule
\begin{align*}
    E(\mu, \sigma) = {\color{red}{}\sigma(\phi(-\frac{\mu}{\sigma}) - \phi(\frac{1-\mu}{\sigma})) +\mu(\Phi(\frac{1-\mu}{\sigma}) - \Phi(-\frac{\mu}{\sigma}))} + {\color{blue}{}1 - \Phi(\frac{1-\mu}{\sigma})}.
\end{align*}
\hrule
\begin{align*}
    V(\mu, \sigma) = &~ \int_{\infty}^{\infty} (\text{ReLU1}(x) - E(\mu, \sigma))^2 \frac{1}{\sigma} \phi(\frac{x - \mu}{\sigma})\,dx \\
    \text{Let }E(\mu, \sigma) = &~\Bar{\mu}. \\
    V(\mu, \sigma) = & \int_{\infty}^{\infty} (\text{ReLU1}(x) - \Bar{\mu})^2 \frac{1}{\sigma} \phi(\frac{x - \mu}{\sigma})\,dx \\
    = & {\color{teal}{}\int_{-\infty}^{0} \Bar{\mu}^2 \frac{1}{\sigma} \phi(\frac{x - \mu}{\sigma})\,dx} + {\color{cyan}{}\int_{0}^{1} (x - \Bar{\mu})^2 \frac{1}{\sigma} \phi(\frac{x - \mu}{\sigma})\,dx} + {\color{violet}{}\int_{1}^{\infty} (1 - \Bar{\mu})^2 \frac{1}{\sigma} \phi(\frac{x - \mu}{\sigma})\,dx}.
\end{align*}
\hrule
\begin{align*}
    {\color{teal}{}\int_{-\infty}^{0} \Bar{\mu}^2 \frac{1}{\sigma} \phi(\frac{x - \mu}{\sigma})\,dx} = &~ \Bar{\mu}^2 \int_{-\infty}^{0} {\color{blue}{}\frac{1}{\sigma} \phi(\frac{x - \mu}{\sigma})\,dx} \\
    = &~{\color{teal}{}\Bar{\mu}^2 \Phi(-\frac{\mu}{\sigma})}.
\end{align*}
\hrule
\begin{align*}
    &{\color{cyan}{}\int_{0}^{1} (x - \Bar{\mu})^2 \frac{1}{\sigma} \phi(\frac{x - \mu}{\sigma})\,dx} \\
    =& \int_{0}^{1} (x^2 - 2\Bar{\mu}x + \Bar{\mu}^2) \frac{1}{\sigma} \phi(\frac{x - \mu}{\sigma})\,dx \\
    =& {\color{gray}{}\int_{0}^{1} \frac{x^2}{\sigma} \phi(\frac{x - \mu}{\sigma})\,dx} + {\color{magenta}{}\int_{0}^{1} - \frac{2\Bar{\mu}x}{\sigma} \phi(\frac{x - \mu}{\sigma})\,dx} + {\color{olive}{}\int_{0}^{1} \frac{\Bar{\mu}^2}{\sigma} \phi(\frac{x - \mu}{\sigma})\,dx}.
\end{align*}
\hrule
\begin{align*}
    {\color{gray}{}\int_{0}^{1} \frac{x^2}{\sigma} \phi(\frac{x - \mu}{\sigma})\,dx} = &~ \frac{1}{\sigma} \int_{0}^{1} x^2 \phi(\frac{x - \mu}{\sigma})\,dx \\
    = & \frac{1}{\sigma} \int_{0}^{1} x^2 \phi(a+bx)\,dx, \text{ with }a = -\frac{\mu}{\sigma} \text{, }b = \frac{1}{\sigma} \\
    \stackrel{\text{Eq.102~\citep*{Owen_1980}}}{=} &~ \left.\frac{1}{\sigma} [\frac{1}{b^3}((a^2 + 1)\Phi(a+bx) + (a-bx)\phi(a+bx))]\right\vert_{0}^{1} \\
    = & \frac{1}{\sigma} \left.[\sigma^3((\frac{\mu^2}{\sigma^2} + 1)\Phi(\frac{x - \mu}{\sigma})
    - \frac{x+\mu}{\sigma} \phi(\frac{x - \mu}{\sigma}))]\right\vert_{0}^{1} \\
    = & \left.\sigma^2((\frac{\mu^2}{\sigma^2} + 1)\Phi(\frac{x - \mu}{\sigma})
    - \frac{x+\mu}{\sigma} \phi(\frac{x - \mu}{\sigma}))\right\vert_{0}^{1} \\
    = &~ \left.(\mu^2 + \sigma^2)\Phi(\frac{x - \mu}{\sigma}) - \sigma(x + \mu) \phi(\frac{x - \mu}{\sigma})\right\vert_{0}^{1} \\
    = &~ (\mu^2 + \sigma^2)\Phi(\frac{1 - \mu}{\sigma}) - (\sigma + \mu\sigma) \phi(\frac{1 - \mu}{\sigma}) - (\mu^2 + \sigma^2)\Phi(-\frac{\mu}{\sigma}) + \sigma\mu \phi(-\frac{\mu}{\sigma}) \\
    = &~ {\color{gray}{}(\mu^2 + \sigma^2)(\Phi(\frac{1 - \mu}{\sigma}) - \Phi(-\frac{\mu}{\sigma}))} {\color{gray}{}- (\sigma + \mu\sigma) \phi(\frac{1 - \mu}{\sigma}) + \sigma\mu \phi(-\frac{\mu}{\sigma})}.
\end{align*}
\hrule
\begin{align*}
    {\color{magenta}{}\int_{0}^{1} -\frac{2\Bar{\mu}x}{\sigma} \phi(\frac{x - \mu}{\sigma})\,dx} = &~ -2\Bar{\mu} (\frac{1}{\sigma} \int_{0}^{1} {\color{red}{}x \phi(\frac{x - \mu}{\sigma})\,dx} ) \\
    = & {\color{magenta}{}-2\Bar{\mu}(\sigma(\phi(-\frac{\mu}{\sigma}) - \phi(\frac{1-\mu}{\sigma})) + \mu(\Phi(\frac{1-\mu}{\sigma}) - \Phi(-\frac{\mu}{\sigma})))}.
\end{align*}
\hrule
\begin{align*}
    {\color{olive}{}\int_{0}^{1} \frac{\Bar{\mu}^2}{\sigma} \phi(\frac{x - \mu}{\sigma})\,dx} = &~ \Bar{\mu}^2 \int_{0}^{1} \frac{1}{\sigma} \phi(\frac{x - \mu}{\sigma})\,dx \\
    = &~ {\color{olive}{}\Bar{\mu}^2(\Phi(\frac{1-\mu}{\sigma}) - \Phi(-\frac{\mu}{\sigma}))}.
\end{align*}
\hrule
\begin{align*}
    &{\color{cyan}{}\int_{0}^{1} (x - \Bar{\mu})^2 \frac{1}{\sigma} \phi(\frac{x - \mu}{\sigma})\,dx} \\
    = &~ {\color{gray}{}(\mu^2 + \sigma^2)(\Phi(\frac{1 - \mu}{\sigma}) - \Phi(-\frac{\mu}{\sigma})) - (\sigma + \mu\sigma) \phi(\frac{1 - \mu}{\sigma}) + \sigma\mu \phi(-\frac{\mu}{\sigma})} \\
    & {\color{magenta}{}-2\Bar{\mu}(\sigma(\phi(-\frac{\mu}{\sigma}) - \phi(\frac{1-\mu}{\sigma})) + \mu(\Phi(\frac{1-\mu}{\sigma}) - \Phi(-\frac{\mu}{\sigma})))} \\
    & {\color{olive}{}+ \Bar{\mu}^2(\Phi(\frac{1-\mu}{\sigma}) - \Phi(-\frac{\mu}{\sigma}))}.
\end{align*}
\hrule
\begin{align*}
    &{\color{violet}{}\int_{1}^{\infty} (1 - \Bar{\mu})^2 \frac{1}{\sigma} \phi(\frac{x - \mu}{\sigma})\,dx} \\
    = &~ (1 - \Bar{\mu})^2 \int_{1}^{\infty} \frac{1}{\sigma} \phi(\frac{x - \mu}{\sigma})\,dx \\
    = &~ {\color{violet}{}(1 - \Bar{\mu})^2 (1 - \Phi(\frac{1-\mu}{\sigma}))}.
\end{align*}
\hrule
\begin{align*}
    V(\mu, \sigma) = & ~ {\color{teal}{}\Bar{\mu}^2 \Phi(-\frac{\mu}{\sigma})} \nonumber \\
    & {\color{gray}{}+ (\mu^2 + \sigma^2)(\Phi(\frac{1 - \mu}{\sigma}) - \Phi(-\frac{\mu}{\sigma})) - (\sigma + \mu\sigma) \phi(\frac{1 - \mu}{\sigma}) + \sigma\mu \phi(-\frac{\mu}{\sigma})} \nonumber \\
    & {\color{magenta}{}-2\Bar{\mu}(\sigma(\phi(-\frac{\mu}{\sigma}) - \phi(\frac{1-\mu}{\sigma})) + \mu(\Phi(\frac{1-\mu}{\sigma}) - \Phi(-\frac{\mu}{\sigma})))} \nonumber \\
    & {\color{olive}{}+ \Bar{\mu}^2(\Phi(\frac{1-\mu}{\sigma}) - \Phi(-\frac{\mu}{\sigma})) + (1 - \Bar{\mu})^2 (1 - \Phi(\frac{1-\mu}{\sigma}))} \nonumber \\
    = &~ (\mu^2 - 2 \mu \Bar{\mu} + \sigma^2 + 2 \Bar{\mu} - 1) \Phi(\frac{1-\mu}{\sigma}) \nonumber \\
    & - (\mu^2 - 2 \mu \Bar{\mu} + \sigma^2) \Phi(-\frac{\mu}{\sigma}) \nonumber \\
    & - (\mu \sigma - 2 \Bar{\mu} \sigma + \sigma) \phi(\frac{1-\mu}{\sigma}) \nonumber \\
    & + (\mu\sigma - 2 \Bar{\mu}\sigma) \phi(-\frac{\mu}{\sigma}) \nonumber \\
    & + \Bar{\mu}^2 - 2 \Bar{\mu} + 1.
\end{align*}

\subsection{Training, Dataset and Implementation Details}\label{sec:training}

\paragraph{Model Architecture.}
For the backbones, we use the pre-trained feature extractors of ResNet50, ResNeXt50, and Wide-ResNet50 provided by torchvision\footnote{torchvision is available at: https://github.com/pytorch/vision}. The resulting number of trainable parameters for \ourmod\ are 24,314,816 (ResNet50), 23,786,688 (ResNeXt50), and 67,641,024 (Wide-ResNet50) for the models trained on CUB-200-2011.
ResNet follows the ResNet18 backbone with just one input channel.
In the original ResNet18, the number of channels is increased every two ResBlocks.
We increase the capacity by adding one additional ResBlock before every increase in the number of channels (the resulting number of trainable parameters for \ourmod\ is 17,825,528).
The ConvNet is the same architecture as our ResNet but without skip connections (the resulting number of trainable parameters for \ourmod\ is 17,319,928).

\paragraph{Data.}
For CUB-200-2011 and Stanford Dogs, we use the official test set as a hold-out test set and randomly split the training set into five folds stratified by the class label.
For RSNA, we use the official training set only.
We split 20\% for a hold-out test set, stratified by age, sex, and class labels.
Additionally, we split the remaining 80\% into five folds.
We resize all images to 224x224 resolution (bilinear interpolation).

\paragraph{Training Details and Evaluation.} For ResNet50, ResNeXt50, and Wide-ResNet50, we initialize weights of the encoder $V$ with pre-trained weights of ImageNet-1k\footnote{ImageNet-1k is available at https://www.image-net.org/download.php} provided by torchvision.
We optimize with AdamW~\citep*{Loshchilov_2017}.
Following the implementation of Chen et al.~(\citeyear{Chen_2019}), we first do a warm-up of parameters of $Z$ and $Q$ for ten epochs, starting at $\frac{1}{10}$-th of the initial learning and linearly increasing every iteration to $\frac{1}{5}$-th of the initial learning rate.
Then, we cycle between optimizing all parameters related to feature extraction ($V, Z, Q$), and classification ($F$) for five and ten epochs, respectively.
During each of these cycles, we perform two learning rate scheduling cycles with the Cyclic Learning Rate Scheduler provided by torchvsision (base learning rate of $\frac{1}{5}$-th of the initial lr, max learning rate is the initial learning rate, linear increase and decrease updated every iteration).
We carry out a grid search for hyper-parameters on one split of CUB-200-2011 with ResNet50 based on the validation balanced accuracy of the validation set of this split.
We follow the training routine proposed in Chen et al.~(\citeyear{Chen_2019}) and set the coefficients before cluster $\mathcal{L}_{\text{clst}}$ and separation cost $\mathcal{L}_{\text{sep}}$ to $\lambda_1 = \lambda_2 = 0.5$:
\begin{align*}
    &\mathcal{L}(y, \mathcal{I}) = \mathcal{L}_{\text{CE}}(y, (F \circ Q \circ Z \circ V)(\mathcal{I})) \\
    &\phantom{\mathcal{L}(y, \mathcal{I}) = } + \lambda_1 \mathcal{L}_{\text{clst}}(y, (Z \circ V)(\mathcal{I})) \\
    &\phantom{\mathcal{L}(y, \mathcal{I}) = } + \lambda_2 \mathcal{L}_{\text{sep}}(y, (Z \circ V)(\mathcal{I})), \\
    &\mathcal{L}_{\text{clst}}(y, (Z \circ V)(\mathcal{I})) =
    \min_{c=y, k, i, j} {\lVert{p^c_k - z_{i,j}}\rVert}^2_2, \\
    &\mathcal{L}_{\text{sep}}(y, (Z \circ V)(\mathcal{I})) = 
    \min_{c\neq y, k, i, j} {\lVert{p^c_k - z_{i,j}}\rVert}^2_2,
\end{align*}
with $\mathcal{L}_{\text{CE}}$ the cross-entropy loss and $y$ the target label of the input image $\mathcal{I}$.
Then, we re-train the model with the best set of hyper-parameters five times (once on each training set of each split) with early stopping for 1000 epochs.
We use this hyper-parameter set to train for all models on CUB-200-2011 and Stanford Dogs.
The hyper-parameter search space is (fat numbers denote the best configuration):
\begin{itemize}
    \item Learning Rate: [0.022, 0.01, 0.0022, 0.001, \textbf{0.00022}, 0.0001]
    \item Weight Decay: [0.0, \textbf{0.001}]
    \item ($K$) Prototypes per Class: [3, \textbf{5}, 7]
    \item ($L$) Channels for Prototypes: [\textbf{128}, 256, 512].
\end{itemize}
We apply random affine transformations (-25..25 degrees, shear of 15, probability of 1.0), followed by random horizontal flip (probability of 0.5), resizing to 224x224 (bilinear interpolation), and normalization (mean=[0.485, 0.456, 0.406], standard deviation=[0.229, 0.224, 0.225]).

For RSNA, we pre-train ConvNet and ResNet backbones five times (once on each training set of each split) to yield five sets of pre-trained weights (one for each split), which prevents bias towards the validation set.
We use a learning rate of 0.001, weight decay of 0.0, and train for 100 epochs.
Then, we carry out a hyper-parameter grid search for ProtoPNet on one split (same search space as for CUB-200-2011) and re-train with the best set of hyper-parameters five times (once on each training set of each split, for 100 epochs with early stopping).

The best configuration for ConvNet is: Learning Rate of 0.0001, Weight Decay of 0.0, ($K$) Prototypes per Class of 7, ($L$) Channels for Prototypes of 256.
The best configuration for ResNet is: Learning Rate of 0.00022, Weight Decay of 0.0, ($K$) Prototypes per Class of 7, ($L$) Channels for Prototypes of 512.
We apply random affine transformations (-45..45 degrees, translation of -0.15..0.15\% of the image size in each direction, the scale of 85..115\%, probability of 1.0), and rescale intensities to 0..1.

As we have five models for each architecture and dataset, we report the mean and standard deviation of the balanced accuracy on the five validation sets and their performance on the test set.
For each model trained on CUB-200-2011 or Stanford Dogs, we randomly select 100 prototypes (total prototypes are 1000 and 600, respectively) and all 14 prototypes for models trained on RSNA.
We create our proposed explanations using 32 network evaluations per input feature for DASP~\citep*{Ancona_2019} (approximately 90 minutes per image on an NVIDIA A10040GB).
We calculate the AOPC score by iterative removal of the most important features, i.e., in every iteration, one more feature (the next, most important one) is replaced with 0.
Therefore, the evaluation of Eq.~10 is expected to decrease faster for higher-quality explanations.
Note that the squared L2-distance has its minimum at 0.
As we evaluate the change in the input image of a prototype, $s_{p^{c}_k}(\mathcal{I}_{\xi{(p_k^{c})}}^{(0)})$ is zero and the whole Eq.~10 becomes negative for all steps.

\subsection{Additional Results}\label{sec:addresults}

We report the Balanced Accuracy (mean and standard deviation in \%) for all models and the datasets on which they were evaluated in Tab.~\ref{tab:clf_full}.
For all models, there is only marginal overfitting towards the validation set (drop-off in performance is $\leq 1\%$ for all models).
Note that ResNeXt50 outperforms Wide-ResNet50 on CUB-200-2011, while it lacks behind 0.6\% on Stanford Dogs.
ConvNet outperforms ResNet on RSNA.
Additionally, the drop-off in performance from the validation set to the test set is 1.4\% for the ConvNet and 1.8\% for the ResNet.
Compared to ResNet, the standard deviation of ConvNet is +5.6\% on the test set, which indicates training instabilities w.r.t. the dataset split.
While it can be argued that ResNet is more stable in this regard, we can observe from Fig.~\ref{fig:rsna_resnet} that ResNet achieved a high test performance by just learning a single prototype for one class.
Therefore, this model should not be applied in clinical practice, which is an important finding only enabled by the architecture of case-based reasoning.

\begin{table*}[!ht]
\centering
\begin{tabular}{llll}
\toprule
      Dataset &        Model Backbone & Validation BAcc &      Test BAcc \\
\midrule
 CUB-200-2011 &     ResNeXt50 &  72.8 $\pm$ 0.5 & 72.0 $\pm$ 0.6 \\
 CUB-200-2011 &      ResNet50 &  70.8 $\pm$ 1.1 & 69.8 $\pm$ 0.8 \\
 CUB-200-2011 & Wide-ResNet50 &  71.0 $\pm$ 0.9 & 70.2 $\pm$ 0.7 \\
Stanford Dogs &     ResNeXt50 &  84.1 $\pm$ 0.3 & 83.4 $\pm$ 0.6 \\
Stanford Dogs &      ResNet50 &  81.0 $\pm$ 0.6 & 80.4 $\pm$ 0.5 \\
Stanford Dogs & Wide-ResNet50 &  84.1 $\pm$ 1.0 & 84.0 $\pm$ 0.6 \\
         RSNA &       ConvNet &  74.0 $\pm$ 10.1 & 72.6 $\pm$ 10.8 \\
         RSNA &        ResNet &  73.7 $\pm$ 4.1 & 71.9 $\pm$ 5.2 \\
\bottomrule
\end{tabular}
\caption{\label{tab:clf_full}Balanced Accuracy (BAcc) scores for the models evaluated in this work.}
\end{table*}

We show the AOPC for all evaluated models in Tab.~\ref{tab:aopc_full} (lower numbers are better).
In all cases, the explanations given by \ourmod\ outperform the explanations given by ProtoPNet by a factor >$10^3$.
Note that the training of ResNet on RSNA collapsed:
The model learned only one prototype for one class and just two prototypes for the other class (see Fig.~\ref{fig:rsna_resnet}).
However, the model still achieved a Balanced Accuracy of 79.8\%.
The even bigger gap between explanations for this model shows a multiplicative effect with respect to duplicate prototypes.

\begin{table*}[!ht]
\centering
\begin{tabular}{lllr}
\toprule
      Dataset &         Model & Explanation &       AOPC($Q \circ Z \circ V$)\\
\midrule
 CUB-200-2011 &     ResNeXt50 &   ProtoPNet &  -0.000446 \\
 CUB-200-2011 &     ResNeXt50 &     \ourmod &  -3.721184 \\
 CUB-200-2011 &      ResNet50 &   ProtoPNet &  -0.000223 \\
 CUB-200-2011 &      ResNet50 &     \ourmod &  -2.848353 \\
 CUB-200-2011 & Wide-ResNet50 &   ProtoPNet &  -0.000834 \\
 CUB-200-2011 & Wide-ResNet50 &     \ourmod &  -3.075180 \\
Stanford Dogs &     ResNeXt50 &   ProtoPNet &  -0.001767 \\
Stanford Dogs &     ResNeXt50 &     \ourmod &  -1.850909 \\
Stanford Dogs &      ResNet50 &   ProtoPNet &  -0.000516 \\
Stanford Dogs &      ResNet50 &     \ourmod &  -3.012445 \\
Stanford Dogs & Wide-ResNet50 &   ProtoPNet &  -0.000717 \\
Stanford Dogs & Wide-ResNet50 &     \ourmod &  -3.006323 \\
         RSNA &       ConvNet &   ProtoPNet &  -0.001360 \\
         RSNA &       ConvNet &     \ourmod &  -8.816450 \\
         RSNA &        ResNet &   ProtoPNet &  -0.000012 \\
         RSNA &        ResNet &     \ourmod & -31.912595 \\
\bottomrule
\end{tabular}
\caption{\label{tab:aopc_full}Area Over the Perturbation Curve (AOPC) scores for the models evaluated in this work.}
\end{table*}

\newpage

\subsection{Additional Visualization of Explanations}\label{sec:morevis}

\begin{figure*}[!ht]
    \centering
    \includegraphics[width=\textwidth]{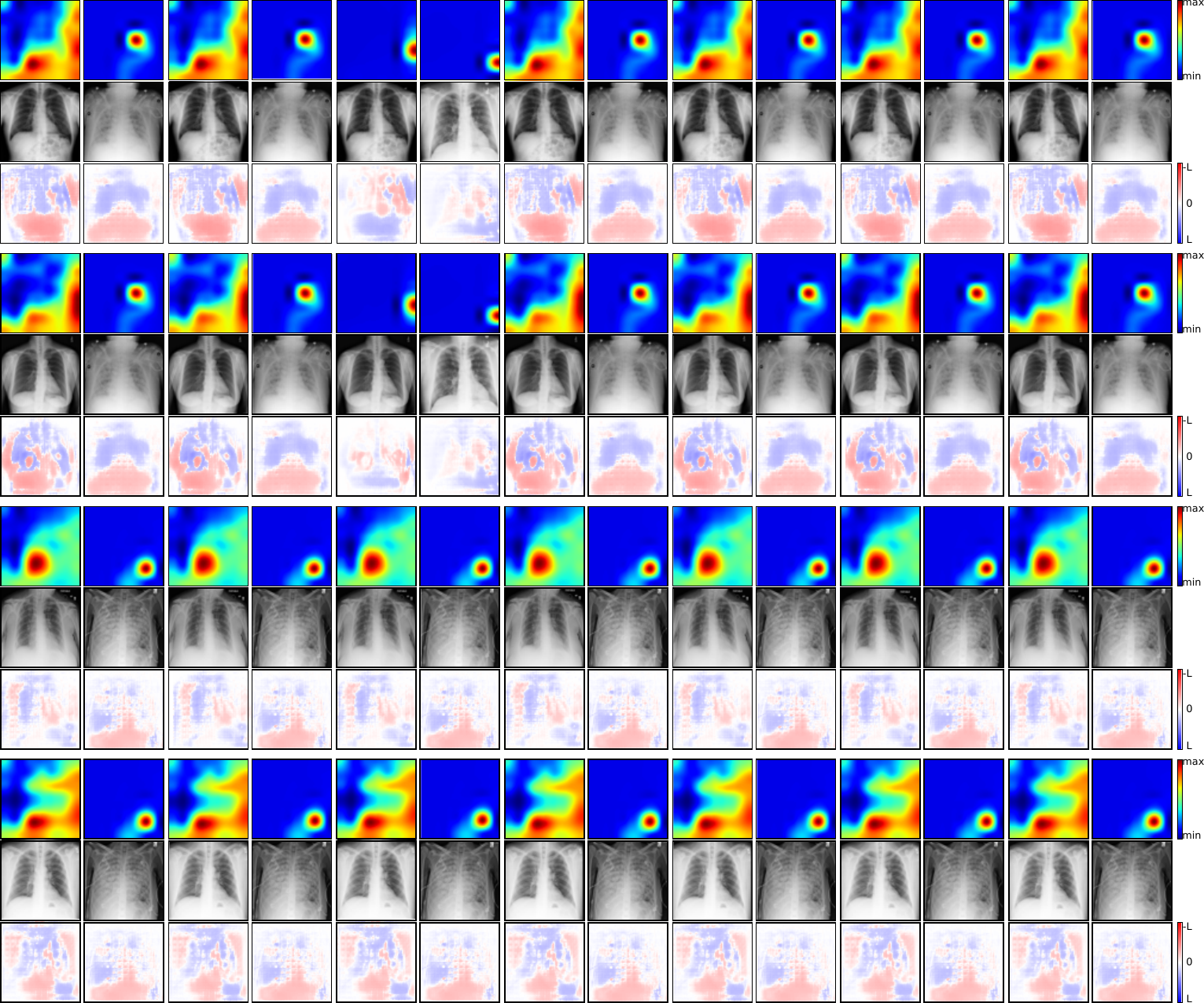}
    \caption{Explanations of the forward pass of a sample with \textbf{ResNet} trained for pneumonia detection on \textbf{RSNA}. We explain the layout of each explanation in the top left with formal notation (left to right, top to bottom): (1) explanation of ProtoPNet for the occurrence of a prototype within the test image; (2) explanation of ProtoPNet for the activation of an image, from which a prototype was extracted; (3) test image; (4) training image, from which the prototype was extracted; (5) explanation of~\ourmod\ for the occurrence of a prototype within the test image; (6) explanation of~\ourmod\ about the image, from which the prototype was extracted.}
    \label{fig:rsna_resnet}
\end{figure*}

\begin{figure*}[!ht]
    \centering
    \includegraphics[width=\textwidth]{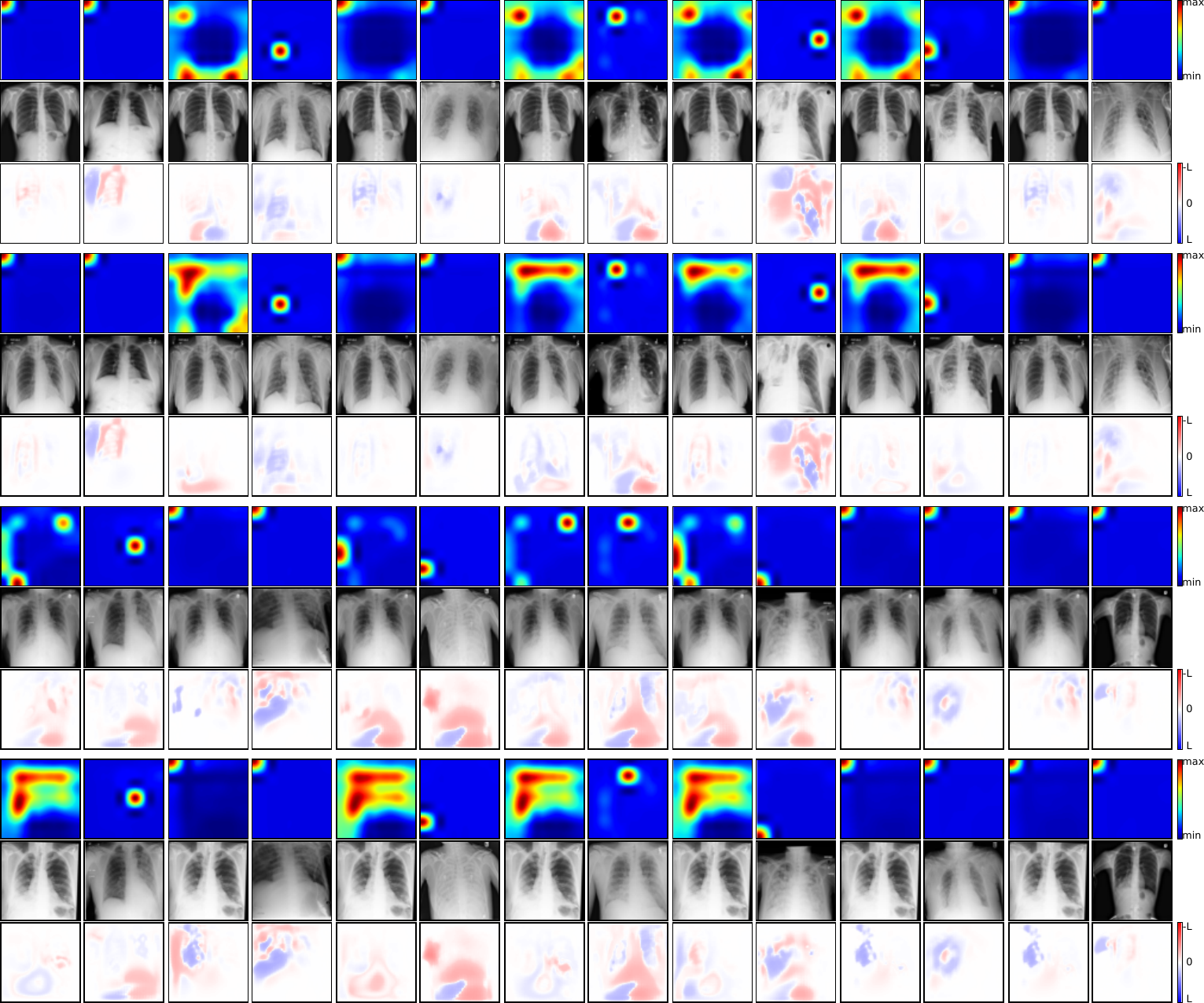}
    \caption{Explanations of the forward pass of a sample with \textbf{ConvNet} trained for pneumonia detection on \textbf{RSNA}. We explain the layout of each explanation in the top left with formal notation (left to right, top to bottom): (1) explanation of ProtoPNet for the occurrence of a prototype within the test image; (2) explanation of ProtoPNet for the activation of an image, from which a prototype was extracted; (3) test image; (4) training image, from which the prototype was extracted; (5) explanation of~\ourmod\ for the occurrence of a prototype within the test image; (6) explanation of~\ourmod\ about the image, from which the prototype was extracted.}
    \label{fig:rsna_convnet}
\end{figure*}

\begin{figure*}[!ht]
    \centering
    \includegraphics[width=\textwidth]{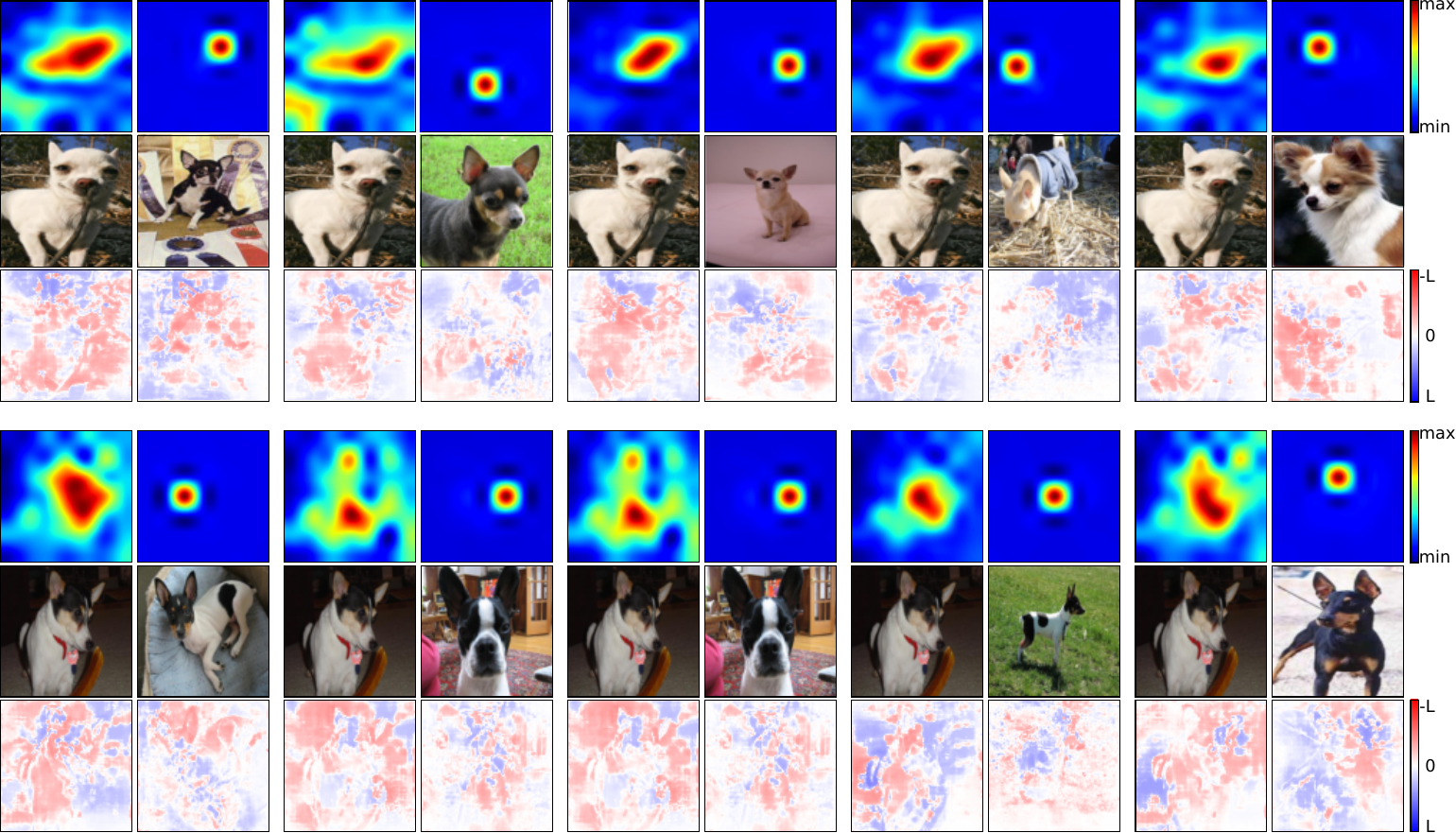}
    \caption{Explanations of the forward pass of a sample with \textbf{Wide-ResNet50} trained for \textbf{dog classification}. We explain the layout of each explanation in the top left with formal notation (left to right, top to bottom): (1) explanation of ProtoPNet for the occurrence of a prototype within the test image; (2) explanation of ProtoPNet for the activation of an image, from which a prototype was extracted; (3) test image; (4) training image, from which the prototype was extracted; (5) explanation of~\ourmod\ for the occurrence of a prototype within the test image; (6) explanation of~\ourmod\ about the image, from which the prototype was extracted.}
    \label{fig:dogs_wideresnet}

\end{figure*}

\begin{figure*}[!ht]
    \centering
    \includegraphics[width=\textwidth]{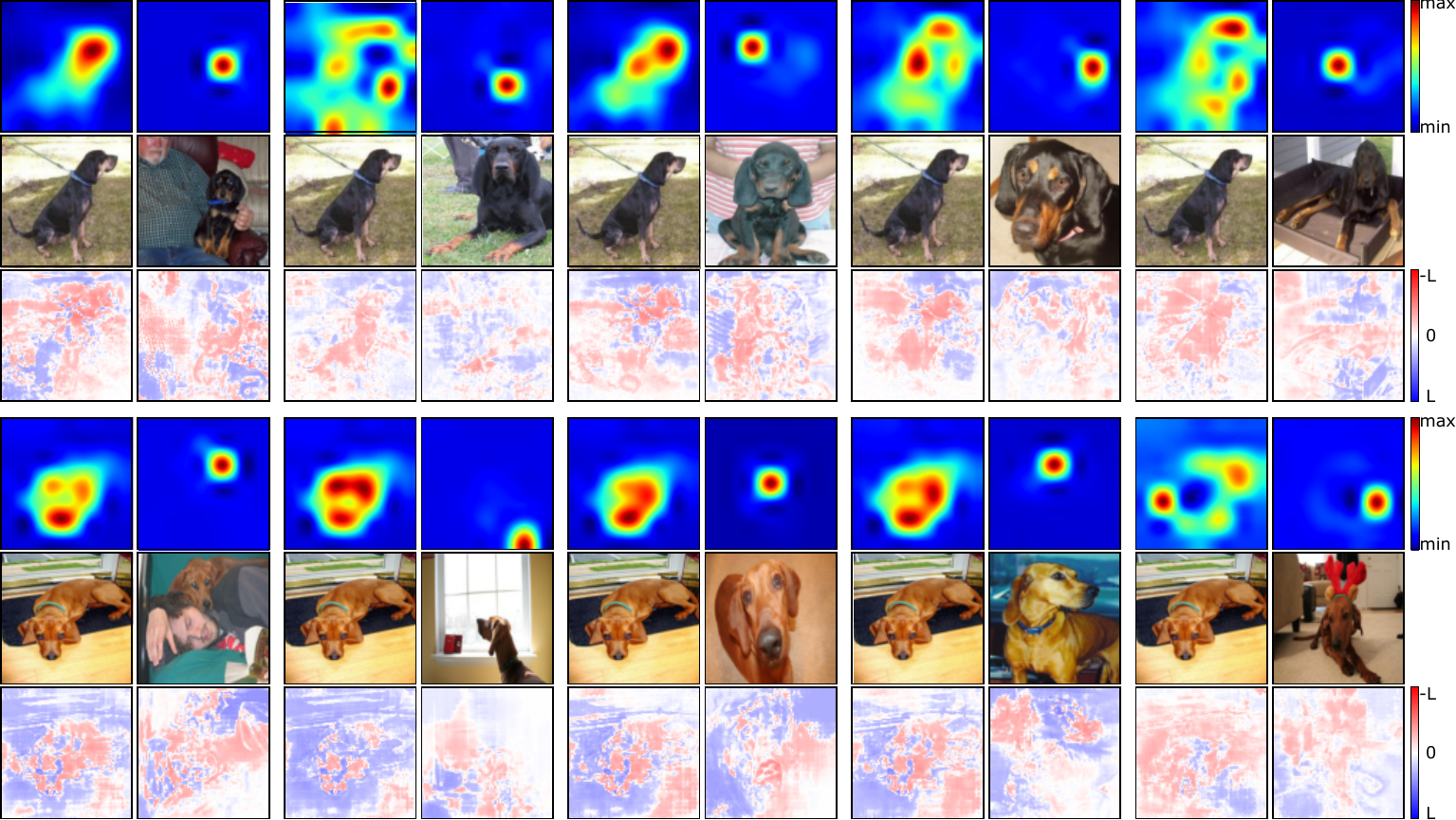}
    \caption{Explanations of the forward pass of a sample with \textbf{ResNeXt50} trained for \textbf{dog classification}. We explain the layout of each explanation in the top left with formal notation (left to right, top to bottom): (1) explanation of ProtoPNet for the occurrence of a prototype within the test image; (2) explanation of ProtoPNet for the activation of an image, from which a prototype was extracted; (3) test image; (4) training image, from which the prototype was extracted; (5) explanation of~\ourmod\ for the occurrence of a prototype within the test image; (6) explanation of~\ourmod\ about the image, from which the prototype was extracted.}
    \label{fig:dogs_resnext}
\end{figure*}

\begin{figure*}[!ht]
    \centering
    \includegraphics[width=\textwidth]{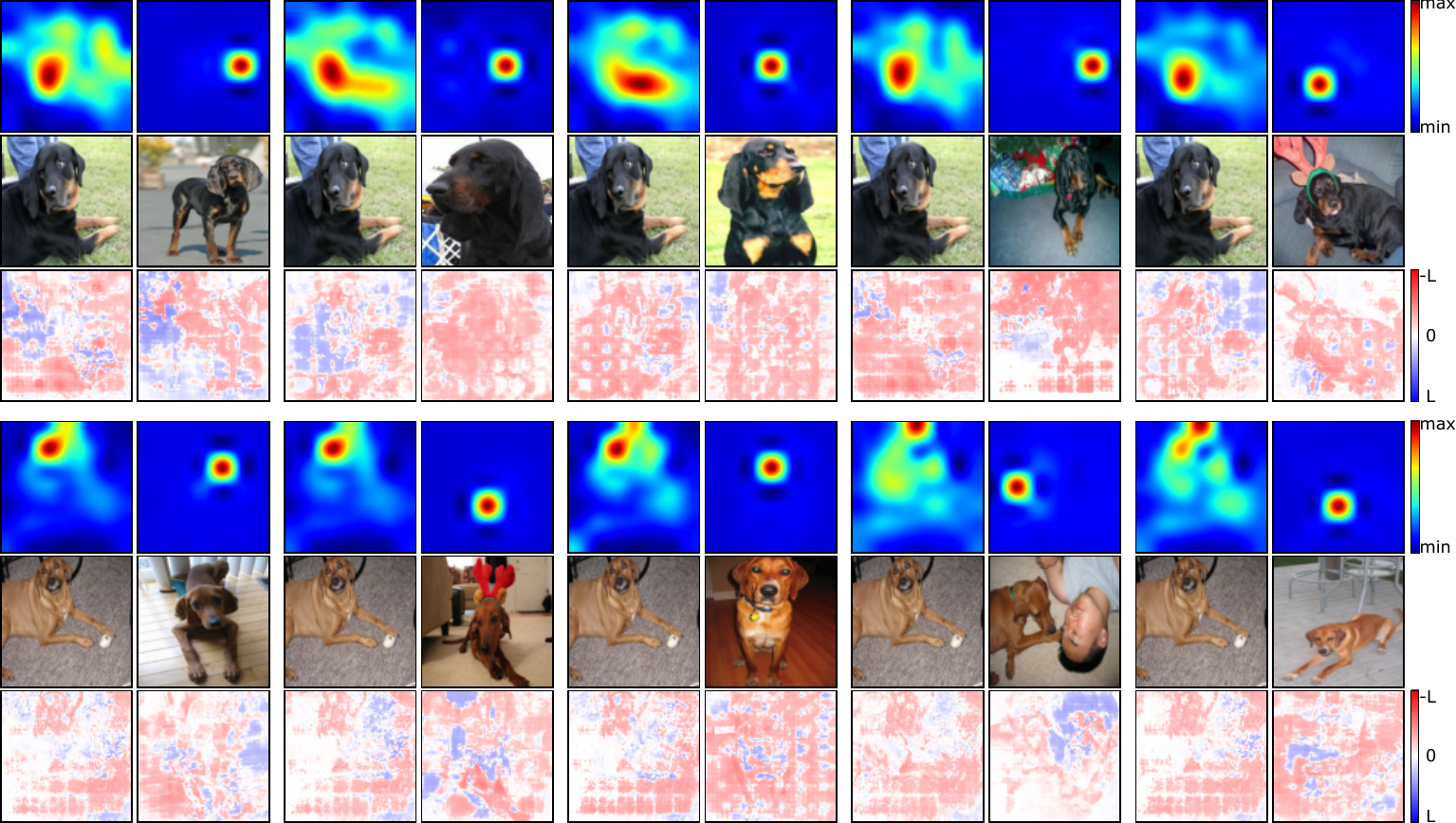}
    \caption{Explanations of the forward pass of a sample with \textbf{ResNet50} trained for \textbf{dog classification}. We explain the layout of each explanation in the top left with formal notation (left to right, top to bottom): (1) explanation of ProtoPNet for the occurrence of a prototype within the test image; (2) explanation of ProtoPNet for the activation of an image, from which a prototype was extracted; (3) test image; (4) training image, from which the prototype was extracted; (5) explanation of~\ourmod\ for the occurrence of a prototype within the test image; (6) explanation of~\ourmod\ about the image, from which the prototype was extracted.}
    \label{fig:dogs_resnet50}
\end{figure*}

\begin{figure*}[!ht]
    \centering
    \includegraphics[width=\textwidth]{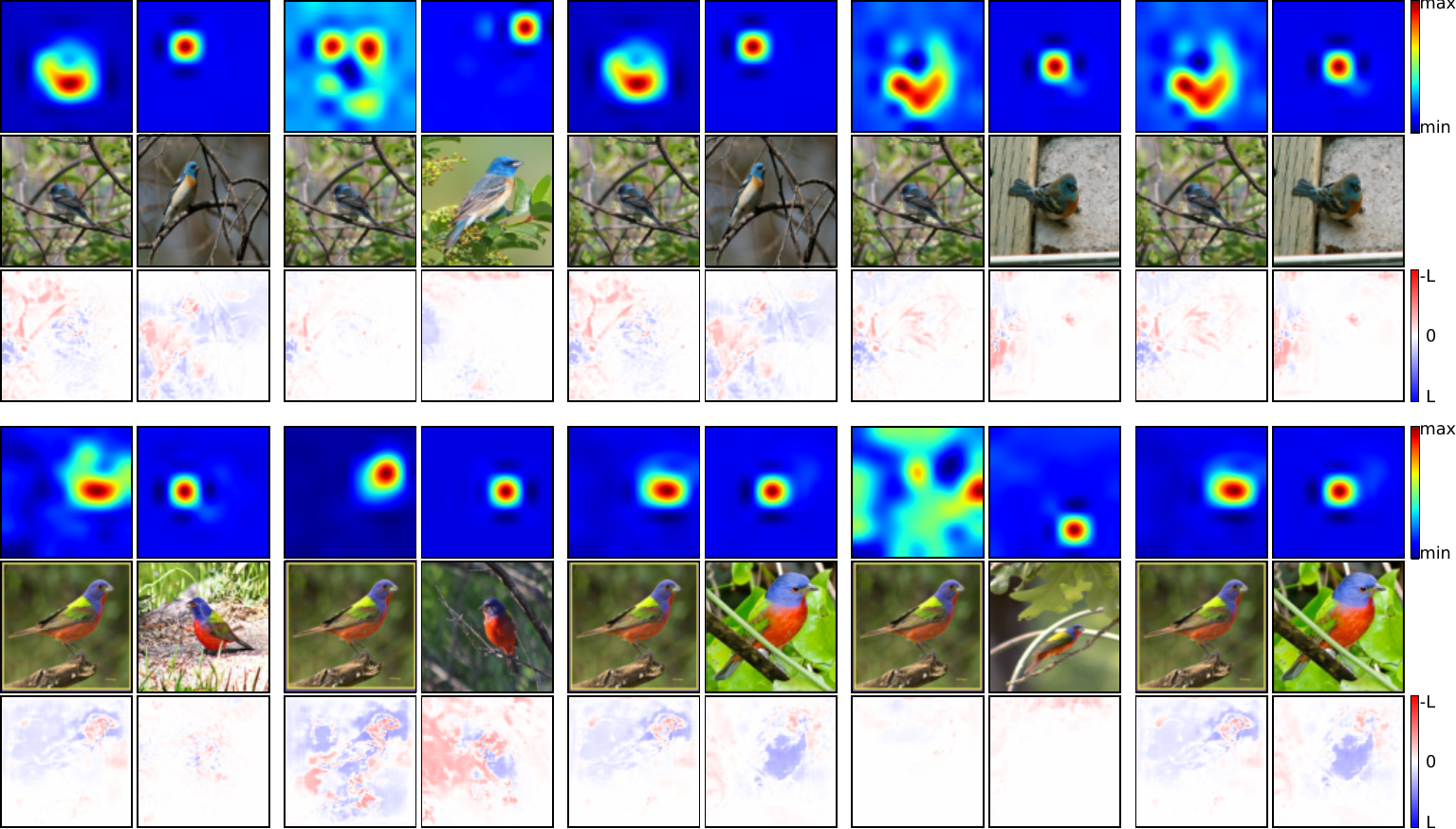}
    \caption{Explanations of the forward pass of a sample with \textbf{Wide-ResNet50} trained for \textbf{bird classification}. We explain the layout of each explanation in the top left with formal notation (left to right, top to bottom): (1) explanation of ProtoPNet for the occurrence of a prototype within the test image; (2) explanation of ProtoPNet for the activation of an image, from which a prototype was extracted; (3) test image; (4) training image, from which the prototype was extracted; (5) explanation of~\ourmod\ for the occurrence of a prototype within the test image; (6) explanation of~\ourmod\ about the image, from which the prototype was extracted.}
    \label{fig:birds_wideresnet}
\end{figure*}

\begin{figure*}[!ht]
    \centering
    \includegraphics[width=\textwidth]{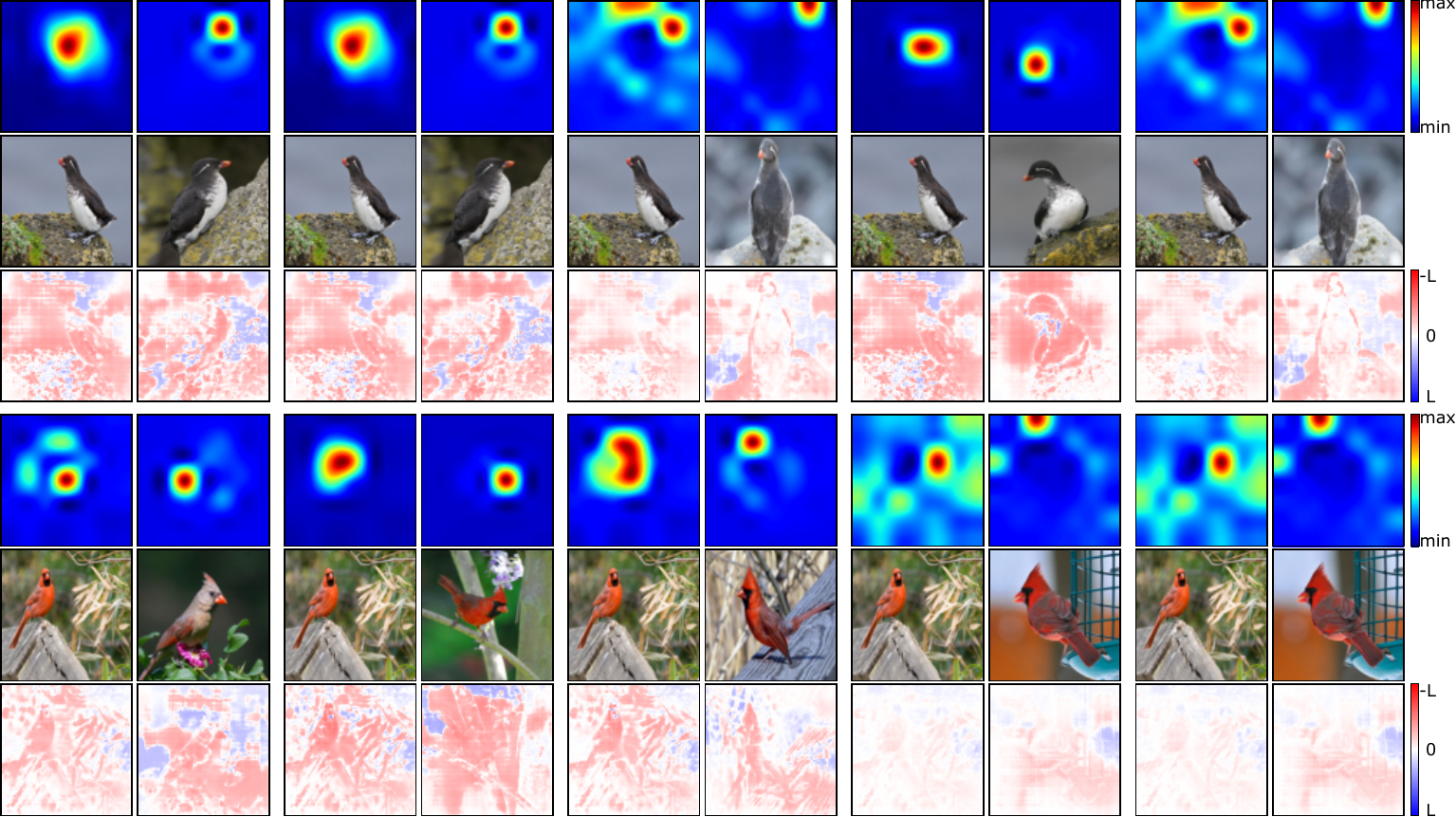}
    \caption{Explanations of the forward pass of a sample with \textbf{ResNeXt50} trained for \textbf{bird classification}. We explain the layout of each explanation in the top left with formal notation (left to right, top to bottom): (1) explanation of ProtoPNet for the occurrence of a prototype within the test image; (2) explanation of ProtoPNet for the activation of an image, from which a prototype was extracted; (3) test image; (4) training image, from which the prototype was extracted; (5) explanation of~\ourmod\ for the occurrence of a prototype within the test image; (6) explanation of~\ourmod\ about the image, from which the prototype was extracted.}
    \label{fig:birds_resnext}
\end{figure*}

\begin{figure*}[!ht]
    \centering
    \includegraphics[width=\textwidth]{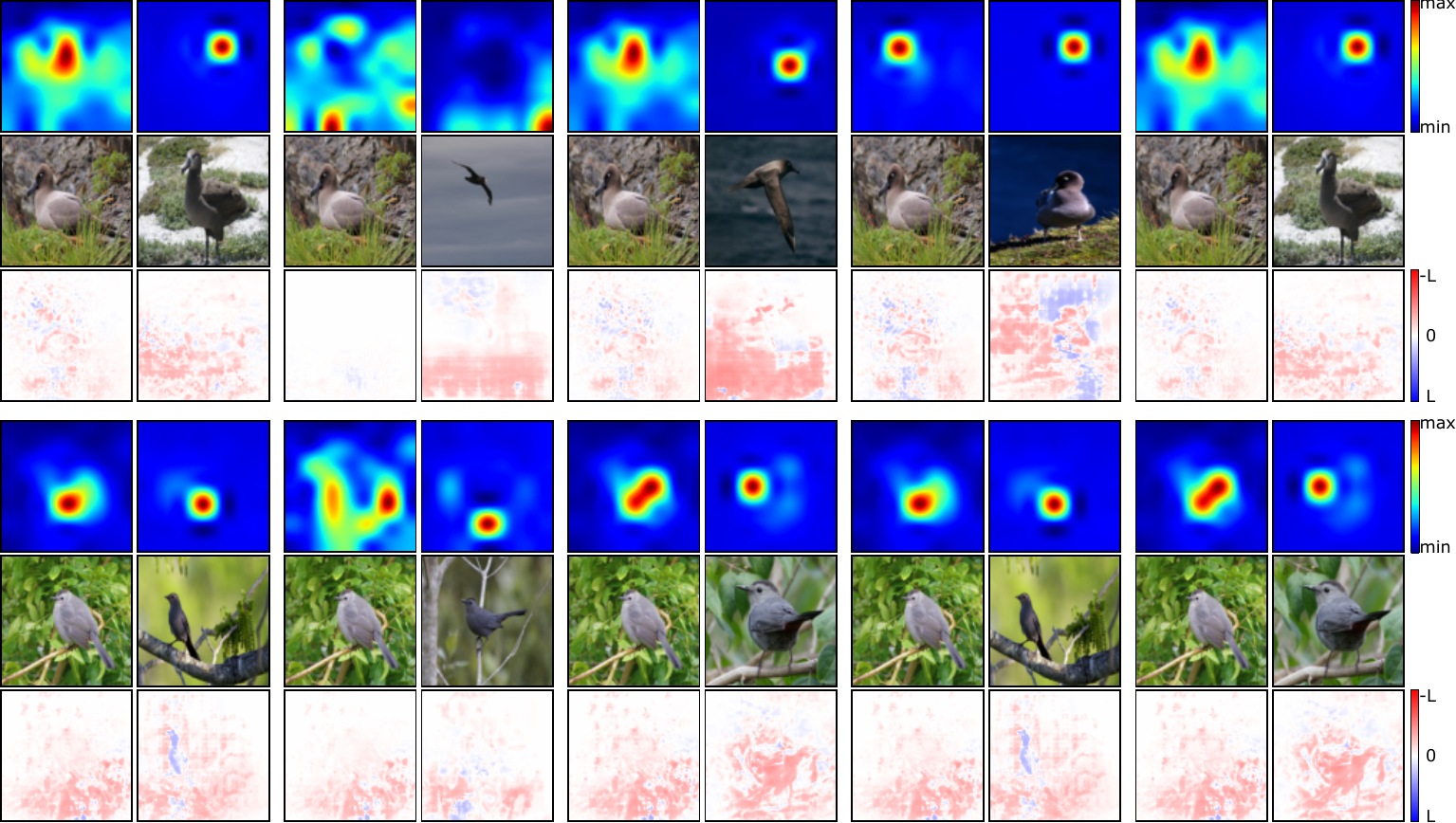}
    \caption{Explanations of the forward pass of a sample with \textbf{ResNet50} trained for \textbf{bird classification}. We explain the layout of each explanation in the top left with formal notation (left to right, top to bottom): (1) explanation of ProtoPNet for the occurrence of a prototype within the test image; (2) explanation of ProtoPNet for the activation of an image, from which a prototype was extracted; (3) test image; (4) training image, from which the prototype was extracted; (5) explanation of~\ourmod\ for the occurrence of a prototype within the test image; (6) explanation of~\ourmod\ about the image, from which the prototype was extracted.}
    \label{fig:birds_resnet50}
\end{figure*}

We perform forward passes for all architectures evaluated in this work and visualize the explanations.
This is a ResNet in Fig.~\ref{fig:rsna_resnet} and a ConvNet in Fig.~\ref{fig:rsna_convnet} on the RSNA dataset.
On the Standford Dogs dataset, this is a Wide-ResNet50 in Fig.~\ref{fig:dogs_wideresnet}, a ResNeXt50 in Fig.~\ref{fig:dogs_resnext}, and a ResNet50 in Fig.~\ref{fig:dogs_resnet50}.
On the CUB-200-2011 dataset, this is a Wide-ResNet50 in Fig.~\ref{fig:birds_wideresnet}, a ResNeXt50 in Fig.~\ref{fig:birds_resnext}, and a ResNet50 in Fig.~\ref{fig:birds_resnet50}.
All figures follow the structure presented in the top left of Fig.~3.

We can see in Fig.~\ref{fig:rsna_resnet} that the prototypes collapsed during training as there are only three prototypes learned in total.
The explanations given by ProtoPNet show high activation for a lot of background of the test sample.
In contrast, the \ourmod\ extracts explanations that lie within the torso.
Additionally, they distinguish between the lung and other parts of the image.
The same accounts for ConvNet trained on RSNA (see Fig.~\ref{fig:rsna_convnet}), where some explanations given by \ourmod\ discard the rips.
In contrast to ResNet, the explanations show that the model learned local features, as most of the input image has little effect on the Shapley value.
However, the explanations given by ProtoPNet are mostly in the background.
For models trained on Stanford Dogs (see Fig.~\ref{fig:dogs_wideresnet}-\ref{fig:dogs_resnet50}), the explanations given by ProtoPNet again suggest locality.
\ourmod\ extracts explanations across the whole input domain, but the contours or body parts of the dogs are clearly visible for most explanations.
On CUB-200-2011 (see Fig.~\ref{fig:birds_wideresnet}-\ref{fig:birds_resnet50}), explanations given by \ourmod\ show a bias towards the background, which is also the case for some explanations of ProtoPNet.
Like the explanations for the models trained on Stanford Dogs, the explanations of \ourmod\ highlight the contours or body parts of birds, like the head, tail, or beak.

In summary, ProtoPNet extracts misleading local explanations, and \ourmod\ extracts granular, faithful explanations.
This prohibits the extraction of crops of the input image for visualization of prototypes, but the explanations of \ourmod\ faithfully show \emph{how} the model transformed the input and \emph{what} lead to the model prediction.
Additionally, granularity showed benefits when applied to medical images.

\clearpage
\bibliographystyle{plainnat}
\bibliography{supplement}